\def\year{2022}\relax

\documentclass[letterpaper]{article} 

\usepackage{aaai23}  
\usepackage{times}  
\usepackage{helvet}  
\usepackage{courier}  
\usepackage[hyphens]{url}  
\usepackage{graphicx} 
\usepackage{natbib}  
\usepackage{caption} 
\DeclareCaptionStyle{ruled}{labelfont=normalfont,labelsep=colon,strut=off} 
\usepackage{algorithm}
\usepackage{algorithmic}
\usepackage{amssymb}
\usepackage{amsthm}
\usepackage{mathtools}
\usepackage{booktabs}
\usepackage{stmaryrd}
\usepackage{multirow}
\usepackage{booktabs}
\usepackage{amssymb}
\usepackage{caption}
\usepackage{subcaption}
\usepackage{xcolor} 
\usepackage{hyperref}

\usepackage{newfloat}
\usepackage{listings}
\DeclareCaptionStyle{ruled}{labelfont=normalfont,labelsep=colon,strut=off} 
\floatstyle{ruled}
\newfloat{listing}{tb}{lst}{}
\floatname{listing}{Listing}
\newtheorem{theorem}{Theorem}
\newtheorem{lemma}[theorem]{Lemma}
\newtheorem*{lemma*}{Lemma}

\newtheorem{proposition}[theorem]{Proposition}
\newtheorem{definition}{Definition}

\newtheorem{assumption}{Assumption}

\usepackage{mathtools}
\newcommand{\defeq}{\vcentcolon=}
\newcommand{\norm}[1]{\left\|{#1}\right\|}
\usepackage{bm}
\newcommand{\xhdr}[1]{{\noindent\bfseries #1}.}

\title{Diffuser: Efficient Transformers with Multi-hop Attention Diffusion\\ for Long Sequences}

\author{
Aosong Feng,
Irene Li,
Yuang Jiang,
Rex Ying
}
\affiliations{

Yale University, New Haven, CT, USA\\
    aosong.feng@yale.edu,
    irene.li@yale.edu, 
    yuang.jiang@yale.edu,
    rex.ying@yale.edu
}

\usepackage{xr-hyper}
\makeatletter
\newcommand*{\addFileDependency}[1]{
  \typeout{(#1)}
  \@addtofilelist{#1}
  \IfFileExists{#1}{}{\typeout{No file #1.}}
}
\makeatother
\newcommand*{\myexternaldocument}[1]{%
    \externaldocument{#1}%
    \addFileDependency{#1.tex}%
    \addFileDependency{#1.aux}%
}
\myexternaldocument{text_appendix}

\usepackage{bibentry}

\def\year{2022}\relax
\usepackage{aaai23}  
\usepackage{times}  
\usepackage{helvet}  
\usepackage{courier}  
\usepackage[hyphens]{url}  
\usepackage{graphicx} 
\usepackage{enumerate}
\usepackage{wrapfig}

\usepackage{natbib}  
\usepackage{caption} 
\DeclareCaptionStyle{ruled}{labelfont=normalfont,labelsep=colon,strut=off} 
\usepackage{algorithm}
\usepackage{algorithmic}
\usepackage{multirow}
\usepackage{booktabs}
\usepackage{amssymb}
\usepackage{amsmath}
\usepackage{xr-hyper}
\usepackage{hyperref}
\usepackage{dsfont}
\usepackage{xcolor}

\usepackage{cleveref}
\usepackage{lipsum}

\crefname{lemma}{Lemma}{Lemmas}

\usepackage{graphics}

\renewcommand{\cite}[1]{\citep{#1}}

\renewcommand{\thesection}{S.\arabic{section}}

\usepackage{mathtools}



\myexternaldocument{text_main}
\usepackage{newfloat}
\usepackage{listings}

\usepackage{amsmath,amsfonts,bm}

\def\rq{{\textnormal{q}}}

\def\ve{{\bm{e}}}

\def\vu{{\bm{u}}}

\def\mA{{\bm{A}}}

\def\mE{{\bm{E}}}

\def\mG{{\bm{G}}}

\def\mW{{\bm{W}}}
\def\mX{{\bm{X}}}

\def\mZ{{\bm{Z}}}

\DeclareMathAlphabet{\mathsfit}{\encodingdefault}{\sfdefault}{m}{sl}
\SetMathAlphabet{\mathsfit}{bold}{\encodingdefault}{\sfdefault}{bx}{n}

\def\sD{{\mathbb{D}}}

\def\sG{{\mathbb{G}}}

\def\sZ{{\mathbb{Z}}}

\def\emZ{{Z}}

\newcommand{\R}{\mathbb{R}}

\newcommand{\mc}[1]{\mathcal{#1}}
\newcommand{\mbf}[1]{\mathbf{#1}}

\newcommand{\ones}{{\mathbf{1}}}

\newcommand{\indic}[1]{\mbf{1}\left\{#1\right\}} 

\newcommand{\reals}{\mathbb{R}} 

\begin{document}

\maketitle

\begin{abstract}
Efficient Transformers have been developed for long sequence modeling, due to their subquadratic memory and time complexity. 
Sparse Transformer is a popular approach to improving the efficiency of Transformers by restricting self-attention to locations specified by the predefined sparse patterns.
However, leveraging sparsity may sacrifice expressiveness compared to full-attention, when important token correlations are multiple hops away.
To combine advantages of both the efficiency of sparse transformer and the expressiveness of full-attention Transformer, we propose \textit{Diffuser}, a new state-of-the-art efficient Transformer.
Diffuser incorporates all token interactions within one attention layer while maintaining low computation and memory costs.
The key idea is to expand the receptive field of sparse attention using \textit{Attention Diffusion},
which computes multi-hop token correlations based on all paths between corresponding disconnected tokens, besides attention among neighboring tokens.
Theoretically, we show the expressiveness of Diffuser as a universal sequence approximator for sequence-to-sequence modeling, and investigate its ability to approximate full-attention by analyzing the graph expander property from the spectral perspective.
Experimentally, we investigate the effectiveness of Diffuser with extensive evaluations, including language modeling, image modeling, and Long Range Arena (LRA). 
Evaluation results show that Diffuser achieves improvements by an average of 0.94\% on text classification tasks and 2.30\% on LRA, with 1.67$\times$ memory savings compared to state-of-the-art benchmarks, which
demonstrates superior performance of Diffuser in both expressiveness and efficiency aspects.
\end{abstract}

\section{Introduction}
Transformers \cite{attention2017vaswani} designed for sequential data have revolutionized the field of Natural Language Processing (NLP) \cite{roberta2019,zhu2020incorporating,li-etal-2022-ligcn}, and have recently made tremendous impact in graph learning \cite{yang2021graphformers,dwivedi2021generalization} and computer vision \cite{dosovitskiy2020image,huynh2022vision}. 
The self-attention used by regular Transformer models comes with a quadratic time
and memory complexity $\mathcal{O}(n^2)$ for input sequence of length $n$, which prevents the application of Transformers to longer sequences in practical settings with limited computational resources.

Recently, many efficient Transformers that improve computational efficiency have emerged.
One line of works approximates the $n\times n$ matrix multiplications by imposing a low-rank assumption on the attention structure,
while the other line of works focuses on sparsification of the attention matrix.
However, the improved computation efficiency always sacrifices expressiveness due to the following challenges: 

\begin{figure}[t]
  \centering
  \includegraphics[width=0.98\linewidth]{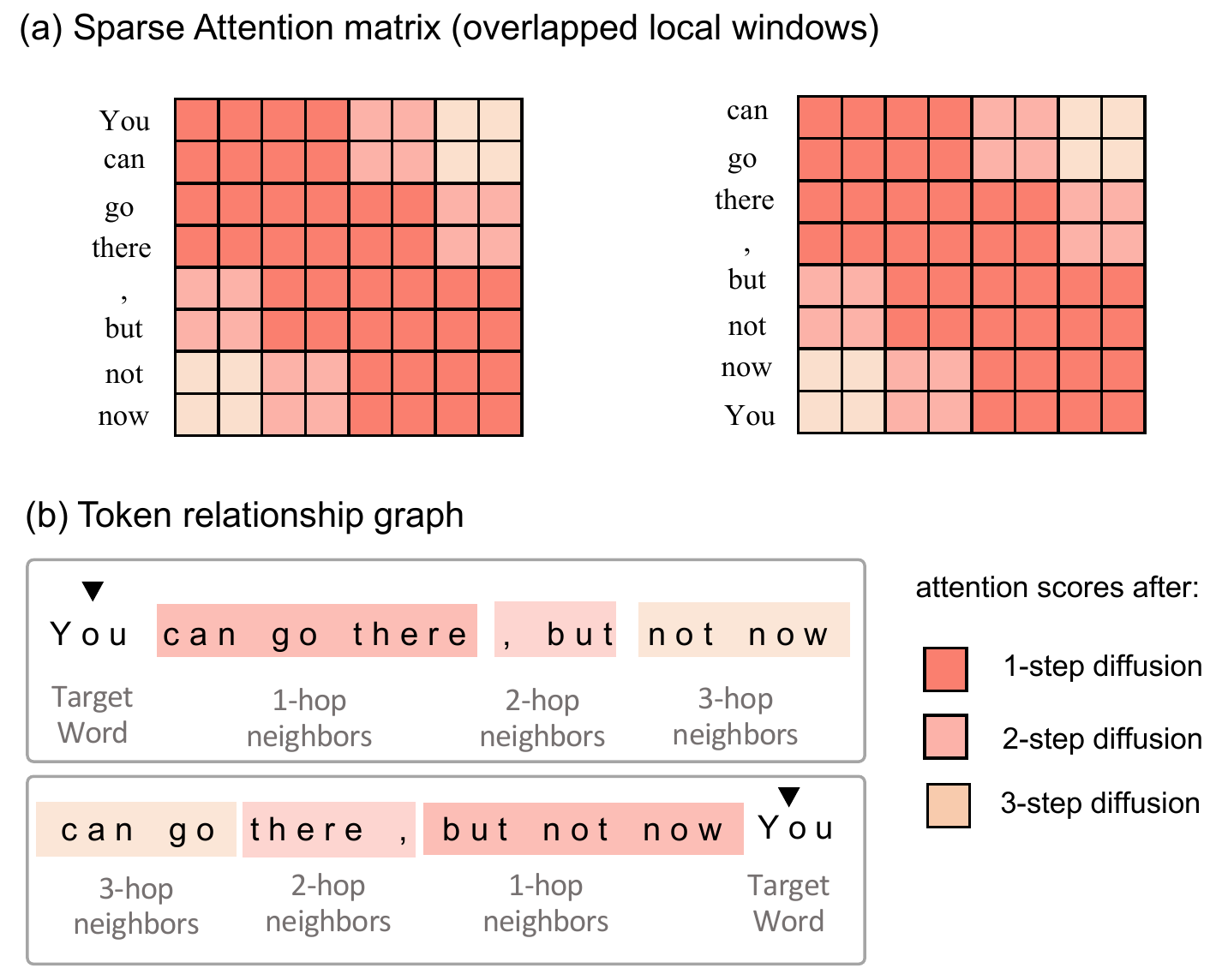}
  \caption{(a) Input token correlations follow predefined sparse pattern (b) The neighborhood structure for the target word completely change by rolling input tokens by 1.}
  \label{fig:example}
  \vspace{-15pt}
\end{figure}

\xhdr{Approximations of full-attention} 
The first line of works avoid explicitly computing $n\times n$ matrix through various approximations such as using the dot-product through kernalization \cite{wang2020linformer,katharopoulos2020transformers} or random projections \cite{peng2021random,choromanski2020rethinking}.
However, such approximations are usually based on strict assumptions about the underlying attention structures such as the low-rank approximation \cite{shen2021efficient,tay2021synthesizer}.
There is currently a lack of rigorous guarantees for these assumptions to hold for potentially full-rank and dense self-attention matrices. Therefore these methods
lead to empirically inferior results in sequence modeling tasks \cite{tay2021long,ren2021combiner}, compared to the sparse Transformer approach. 

\xhdr{Slow information propagation} 
The current state-of-the-art uses the sparse Transformer approach to approximate the full self-attention (see a local pattern example in Figure 1)\cite{zaheer2020bigbird}.
However, such sparsity-based approach can be lossy or even misleading in capturing important token correlations when they are not directly connected.
For example, as the sentence shown in Figure \ref{fig:example}, one-hop attention scores of every pair of neighboring tokens can be misleading (caused by the word \lq\lq can\rq\rq) when the real important correlations (the word \lq\lq but\rq\rq, and \lq\lq not\rq\rq ) are two or three hops away.
A sparse-attention layer only focuses on neighboring tokens, resulting in slower information propagation in the attention graph.
Consequently, to model these crucial long-range correlations, sparse Transformers require more layers to expand the receptive field compared to full-attentions \cite{child2019generating,ho2019axial,dai2019Transformer}.
Some existing works \cite{ainslie2020etc,beltagy2020longformer} deal with the slow propagation by introducing global attentions for important tokens, which alleviates the problem of long-range interactions.
However, we cannot soly rely on global tokens for such propagation because of information loss when aggregating all tokens.

\xhdr{Robustness to input perturbations}
Since attention graph is built upon predefined topology pattern, the attention process can become very different even with minor input sequence changes.
As shown in Figure \ref{fig:example}, being shifted by one token, the attention structure and neighboring tokens of the target word will completely change, leading to inconsistent outputs.
Compared to full-attention where every token is attended regardless of its position, sparse Transformers are less robust to such input perturbations.
Slow information propagation of sparse Transformers will amplify such attention inconsistency, as the inconsistency accumulates when the attention receptive fields gradually expand.

\xhdr{Proposed work} To address the mentioned expressiveness issues and further improve Transformer efficiency, we propose \textit{Diffuser}, a novel sparse Transformer that achieves state-of-the-art performance on sequence modeling with 1.67$\times$ memory savings compared to state-of-the-art efficient Transformers.
The key insight is to introduce \textit{Attention Diffusion} mechanism based on the designed sparse pattern for enabling efficient full-attention and larger receptive field.
Diffuser first calculates attention scores on edges of the attention graph as in most sparse Transformers, 
then computes attention scores between other node pairs through the attention diffusion process. 
Unlike all existing sparse Transformers, Diffuser can model correlations among all pairs of tokens in a single Transformer layer, which extends the attention receptive field to the entire sequence, with minimal runtime overhead.
Moreover, Diffuser designs sparse patterns with better graph expander properties which can be further improved by the attention diffusion.
We theoretically show that Diffuser can be more efficient (requires fewer layers) universal approximators for sequence modeling compared to all existing sparse Transformers, and 
has good properties to approximate the full-attention.

We further demonstrate the performance of Diffuser with datasets from various domains, including text and images, following the standard pretraining and  finetuning procedure. Experiments demonstrate Diffuser's superior performance in expressiveness and efficiency. Compared with state-of-the-art efficient Transformers, Diffuser improves state-of-the-art by an average of 0.94\% on text classification and 2.30\% on LRA for long  sequence modeling, with 1.67$\times$ memory savings and comparable running time. Furthermore, Diffuser achieves state-of-the-art on 2 questions answering tasks and 2 image density estimation tasks. 
The implementation can be found at \url{https://github.com/asFeng/Diffuser}.

\section{Related Work}

\xhdr{Efficient Transformers} 
Many works aim to optimize Transformers for longer inputs. Notably, Bigbird \cite{zaheer2020bigbird} introduced a sparse attention method that considers random, windowed, and global attention, improving performance on tasks including question answering and summarization. Similarly, Longformer \cite{beltagy2020longformer} presented a combination of windowed self-attention and global attention to sparsify the dense attention. Sparse sinkhorn attention \cite{tay2020sparse} and Reformer \cite{kitaev2020reformer} adopted learnable patterns on the attention module. \citet{vyas2020fast} proposed clustered attention that computes attention for only the centroids in clustered queries. Other works focus on kernel-based and feature mapping methods, like Performer \cite{marcin2021performer}, Reformer \cite{kitaev2020reformer} and Linformer \cite{wang2020linformer}. 
Such methods improve self-attention efficiency by grouping, clustering or designing fix sparse patterns, at the expense of expressiveness. In contrast, Diffuser approximates full-attention using attention diffusion on a new sparse pattern, backed by a novel theoretically guaranteed graph expander perspective. 

\xhdr{Diffusion on Graphs}
In graph neural networks (GNNs), it is possible to increase number of layers to facilitate interactions with neighbors that are multiple hops away, but such indirect communication is less effective due to GNN aggregations and results in an increased computational cost. Another solution is to apply diffusion in each graph layer considering the multi-hop neighborhood \cite{graphheat,DCNN}.
\cite{klicpera2019predict} proposed 
PPNP that applies personalized PageRank to propagate node predictions. 
\cite{wallach2019diffusion} propose 
GDC to allow propagation of multi-hop neighbors with generalized graph diffusion. Moreover, \citet{wang2021multi} proposed 
MAGNA, which applies a diffusion based on the attention values in graph attention. Diffuser is inspired by the successful practice of diffusion in the graph domain, and utilizes it to improve the sparse Transformer expressiveness for general sequence modeling.

\section{Diffuser: Multi-hop Attention with Diffusion}
In this section, we define the attention diffusion process and introduce the Diffuser model by integrating the attention diffusion into Transformers with sparse attention patterns.

\subsection{Preliminaries}
\xhdr{Multi-head Self-attention}
Transformers and self-attention mechanism \cite{attention2017vaswani} are proposed for modeling sequences. The input sequence to the $l$-th layer with $n$ tokens can be denoted as $H^{(l-1)}=[x_1,x_2,...,x_n]$, where $H^{(l-1)}\in \mathbb{R}^{n\times d}$, and each token $x_i$ is a $d$ dimensional vector. The attention mechanism introduces matrices $Q, K, V$ as \textit{queries}, \textit{keys}, and \textit{values}, which are linear projections of the input sequences:
\begin{equation}
\label{eq:qkv}
\begin{split}
   Q = XW_Q, K = XW_K, V=XW_V.
\end{split}
\end{equation}

The attention matrix $A$ among tokens is then calculated as the scaled dot-product of queries and keys, and is used to calculate the updated token values:
\begin{equation}
\label{eq:softmax}
\begin{split}
    Attn(X)=AV, A=\mathbf{softmax}\left(\dfrac{QK^\intercal}{\sqrt{d}}\right),
\end{split}
\end{equation}
where $\text{softmax}$ denotes the row-wise softmax normalization, and we omit the bias term for simplicity.
To allow a token to attend to multiple aspects, self-attention can be further extended to multi-head self-attention as follows:
\begin{equation}
\label{eq:msa}
\begin{split}
   M\text{-}Attn(X) = \mathbf{cat}[Attn(X)_1,...,Attn(X)_h]W_O,
\end{split}
\end{equation}
where $h$ is the number of heads in use.

\xhdr{Sparse attention}
The runtime bottleneck of the standard self-attention is the attention matrix $A$ with shape $n$$\times$$n$ in Equation \ref{eq:softmax}, which has to be fully materialized in memory and scales quadratically as input length.
This is impractical for long sequences with large $n$.
To avoid the increased memory usage and speed up the attention calculations, we define the \textit{sparse attention mechanism} described by a directed attention graph $G\!=\!(\mathcal{V},\mathcal{E})$. In this graph, we have tokens to be the nodes, $\mathcal{V}\!=\! \{1,...,n\}$, with the corresponding adjacency matrix $A\!\in\!\{0,1\}^{n\times n}$. Each edge in the graph represents the query-key pair which we will include during computing sparse attention, i.e., $A_{i,j}=1$ if query $i$ attends to key $j$ and is zero otherwise. 
Matrix $A$ can be seen as a mask applied to the full-attention matrix by element-wise multiplication.
The resulting sparse self-attention mechanism in Equation \ref{eq:softmax} can then be rewritten in the token-wise form as
\begin{equation}
\label{eq:sparse_att}
\begin{split}
    Attn(x_i)=\mathbf{softmax}\left( \dfrac{Q_iK_{Ne(i)}^\intercal}{\sqrt{d}}\right)V_{Ne(i)},
\end{split}
\end{equation}
where $x_i$ is the $i$-th input token to update value and $Ne(i)$ represents the neighbors of token $i$ in the attention graph $G$. 

\subsection{Transformer Attention Diffusion}
Similar to other sparse Transformers, the attention matrix $A$ is first calculated on edges of the underlying graph $G$ which is used to characterize the interaction strength between neighboring nodes on the graph, i.e.,
\begin{equation}
\label{eq:sparse_att}
\begin{split}
    A_{i,j}=\dfrac{exp(Q_iK_j/\sqrt{d})}{\sum_{j\in Ne(i)}exp(Q_iK_j/\sqrt{d})  }.
\end{split}
\end{equation}
Each entry of attention matrix $A$ is the attention score between 1-hop neighbors of $G$. Such 1-hop correlations in sparse Transformers cannot include all possible correlations compared to full-attention, which leads to limitations to capturing important correlations when the true dependencies are in several-hops away and not directly connected by edges in the graph as discussed in Figure \ref{fig:example}. 

The key idea of Diffuser is to apply the attention diffusion mechanism to calculate the multi-hop token relationships on the attention graph based on attention weights on edges. 
The multi-hop attention scores are calculated as entries of the graph diffusion matrix $\mathcal{A}$:
\begin{equation}
\label{eq:sparse_att}
\begin{split}
    \mathcal{A} = \sum_{k=0}^{\infty} \theta_{k}A^{k},
\end{split}
\end{equation}
where $A$ is the adjacency matrix or calculated sparse attention matrix, and the weighting coefficient $\theta_k$ satisfies $\sum_{k=0}^{\infty}\theta_k=1, \theta_k \in [0,1]$. The original receptive fields defined by the sparse attention pattern will be gradually expanded as $k$ becomes larger. 
The resulting attention score $\mathcal{A}_{i,j}$ incorporates all paths between token $i$ and $j$, weighted by the coefficient $\theta_k$. 
We then multiply each value vector $V$ by the diffusion attention matrix $\mathcal{A}$, which is equivalent to the message aggregation step in GNN.

Computing the power of attention matrices in Equation \ref{eq:sparse_att} can be inevitably expensive for long sequences, even when the sparsity is considered.
To efficiently apply the diffusion mechanism in Transformers, we implement the graph diffusion process as Personalized PageRank (PPR) by specifying $\theta_k=\alpha(1-\alpha)^k$ with teleport probability $\alpha$. The resulting diffusion matrix $\mathcal{A} = \sum_{k=0}^{\infty} \alpha(1-\alpha)^kA^{k}$ is the power expansion of the solution to the recursive equation $\mathcal{A}=\alpha I+(1-\alpha)\mathcal{A} A$. We then adopt the power iteration method \cite{page1999pagerank} to achieve linear computational complexity by approximating PPR within the the first $K$ diffusion steps. Each power iteration (diffusion) step is calculated as 
\begin{equation}
\label{eq:propagate}
\begin{split}
    Z_{(0)}=V=XW_V,\
    Z_{(k+1)}=(1-\alpha)AZ_{(k)}+\alpha V,
\end{split}
\end{equation}
for $0\leq k<K$. $Z_{K}$ is output of the attention diffusion process, and will converge to the real output $\mathcal{A}V$ as $K\shortrightarrow\infty$ (shown in Appendix \ref{apsec:diff}).

\subsection{Sparse Pattern Design}
\begin{figure}[ht]
  \centering \includegraphics[width=0.98\linewidth]{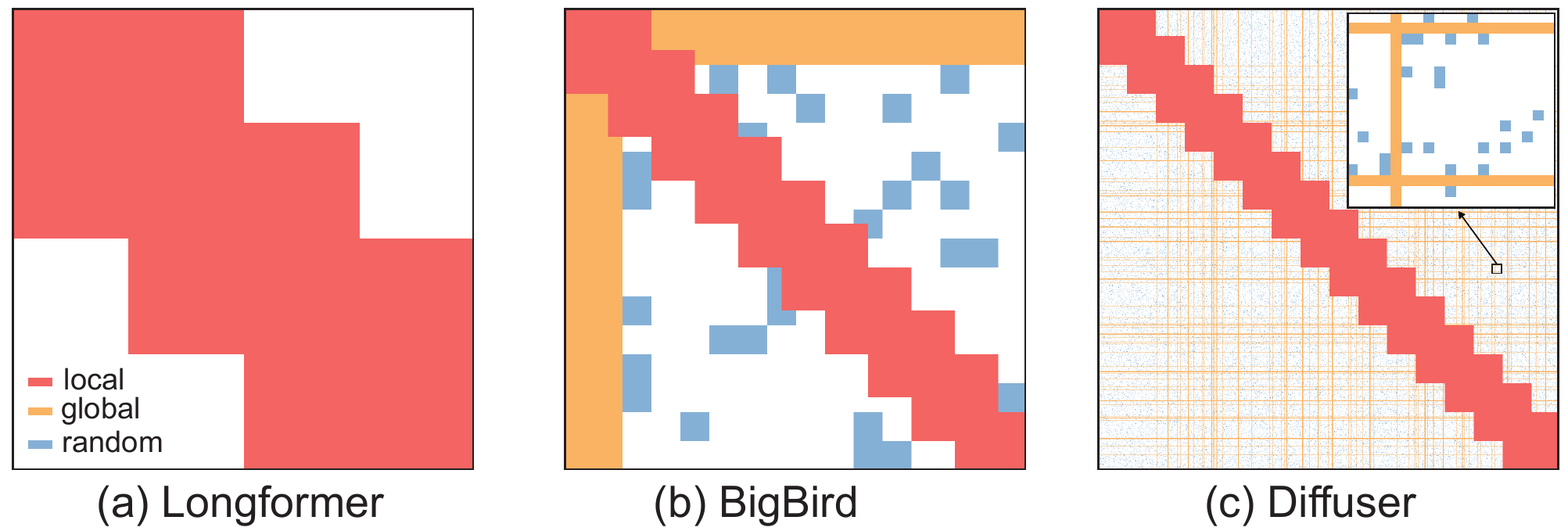}
  \caption{Comparison of sparse patterns (1024$\times$1024) with different types of attentions.}
  \label{fig:pattern}
\end{figure}
Another important ingredient of Diffuser is the design of sparse attention pattern. 
It should be noted that attention diffusion is compatible with any sparse patterns. We design new sparse patterns to leverage the advantages of attention diffusion while maintaining computational efficiency.
As shown in Figure \ref{fig:pattern}, we consider a combination of local window attention, global attention, and random attention to capture token interactions without quadratic complexity dependency on the sequence length.

\xhdr{Local window attention} Local window attentions are constructed by the sliding window and are proposed to model the information locality %
among neighboring tokens, e.g., the proximity of tokens in linguistic structure and the clustering coefficient in the graph.
Given a fixed window size $w$, each token attends to $\frac{1}{2}w$ tokens on each side, and we also consider the cross-window attention by overlapping $\frac{1}{2}w$ tokens, resulting in computational complexity $\mathcal{O}(nw^2)$ which scales linearly with input sequence length $n$.
The resulting receptive field (keys that each query looks up to calculate self-attention) is expanded linearly with more diffusion steps and Transformer layers. For example, the size of the receptive field grows as $(\frac{1}{2}+k)w$ with diffusion step $k$.
Compared to Longformer and BigBird, Diffuser can achieve good expressiveness with smaller local window and therefore sparser attention, because of the fast receptive field expansion by attention diffusion given the same number of attention layers.

\xhdr{Global attention} We introduce global attention by extending the receptive field of tokens to the entire input sequence. Specifically, we randomly choose $g$ tokens among input sequence as global tokens, such that for any global token $i$, $A_{i,:\cdot }=1$ and $A_{\cdot :,i}=1$, resulting in complexity $\mathcal{O}(gn)$. Global tokens share the same set of weight parameters with other types of attentions (in contrast to different weights used in Longformer). 
Furthermore, compared to BigBird which selects global attentions by grouping adjacent tokens, Diffuser constructs global attention with the unit of individual tokens. 

\xhdr{Random attention}
We consider adding random attentions to accelerate the information flow between any pair of nodes.
The intuition of introducing random attention is to enhance the graph expander properties for better full-attention approximation.
From the graph theory perspective, random graph, e.g., Erdős–Rényi graph \cite{erdHos1960evolution}, has been shown to have good expander properties to approximate the complete graph (full-attention) spectrally (detailed in the next section). 
Therefore, for each input token $i$, we randomly select $r$ tokens ($r\ll n$ and above the threshold $\mathcal{O}(log(n)/n)$), such that $A_{i,j}=1$ for each selected token $j$, resulting in a total number of $\mathcal{O}(rn)$ random attentions. 
Compared to BigBird whose random attentions are based on the unit of blocks (e.g., 64 adjacent tokens as a block), Diffuser constructs random attention with the unit of individual tokens. 
Given the same number of global and random attention budget, the token-wise selection leads to more uniform attention distributions with weaker clustering, compared to block-wise selections, which improves the expander properties and accelerates the attention flows among tokens. 
It is noted that the reason BigBird adopts block-wise attentions is to blockify lookups for efficient implementations of attention calculations. In comparison, we implement token-wise attention using commercial graph packages with optimized GSpMM kernels and achieve similar efficiency. 

\begin{table}[h]
\centering
\small
\resizebox{0.98\columnwidth}{!}{
\begin{tabular}{c|c|c|c|c|c|c}
 \hline
\multicolumn{1}{c|}{\multirow{2}{*}{Length}} & \multicolumn{1}{c|}{\multirow{2}{*}{\textbf{Longformer}}} & \multicolumn{1}{c|}{\multirow{2}{*}{\textbf{BigBird}}} & \multicolumn{4}{c}{\textbf{Diffuser}}                                                                          \\ \cline{4-7}  
\multicolumn{1}{c|}{}                        & \multicolumn{1}{c|}{}                            & \multicolumn{1}{c|}{}                         & \multicolumn{1}{c|}{tot} & \multicolumn{1}{c|}{loc} & \multicolumn{1}{c|}{glob} & \multicolumn{1}{c}{rand} \\

 \hline
1024 & 62.5       & 55.7             & \textbf{24.0}        &18.0 &4.2 &1.9     \\
2048 & 34.4       & 32.5             & \textbf{15.5}        &9.2  &4.2 &2.1     \\
4096 & 18.0       & 16.9             & \textbf{11.2}        &4.6  &4.3 &2.2     \\

 \hline
\end{tabular}
}
\caption{The percentage of attentions (non-zero entries in the sparse pattern) with different input lengths. \texttt{tot}, \texttt{loc}, \texttt{glob}, and \texttt{rand} represent the percentage of total, local, global, and random attentions, respectively. Detailed statistics for each type of attention is shown in Appendix \ref{apsec:sparsity}}
\label{tab:pattern}
\end{table}

\subsection{Model Architecture}
\begin{figure}[t]
\centering
  \includegraphics[width=0.82\linewidth]{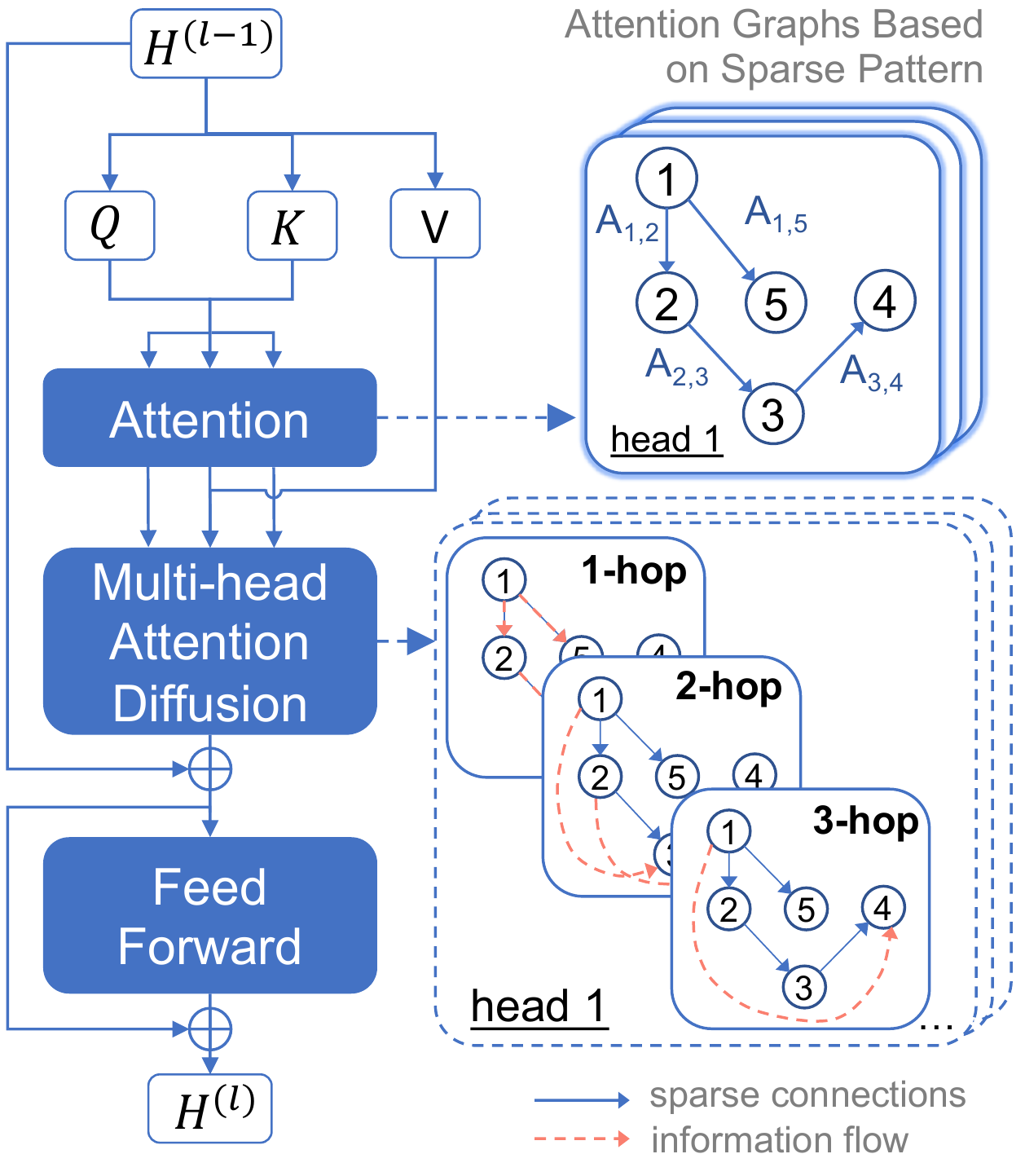}
  \caption{The layer architecture of Diffuser.  }
  \label{fig:archi}
  \vspace{-10pt}
\end{figure}
We introduce the building block of Diffuser, based on the proposed sparse pattern, regular self-attention mechanism, and attention diffusion process, as shown in Figure \ref{fig:archi}.
At layer $l-1$, input $H^{(l-1)}$ is mapped to queries, keys, and values, and attention scores are calculated on edges of the predefined attention graph using scaled dot-product.
Then, attention diffusion procedure is calculated following Equation \ref{eq:propagate} up to $K$ diffusion steps, which spreads the information of tokens to its multi-hop neighbors. 
The residual and feed-forward layers are then used to output $H^{(l)}$ for skip-connections and value mappings.

\section{Theoretical Expressive Power of Diffuser }
In this section, we investigate the expressiveness of the proposed model from two perspectives. 
First, we show Diffuser's capability to \textit{approximate sequence modeling} by proving that the model with sparse connections and diffusion is a universal approximator to sequence-to-sequence function, and it requires less layers to achieve the same expressivity compared with sparse attentions without diffusion.
Second, we show Diffuser's capability to \textit{approximate full-attention}. From the spectral graph perspective, we show that the proposed sparse patterns combined with diffusion induces better graph expander properties, enabling approximations of the complete graph. 

\subsection{Diffuser as Universal Approximators}
We follow the proof of  \citet{yun2019transformers}
and show Diffuser can approximate arbitrary sequence-to-sequence functions (mapping sequential input $X$ from $\mathbb{R}^{n\times d}$ to $\mathbb{R}^{n\times d}$).
Given one family of Diffuser structure $\mathcal{D}^{h,m,r}$ with $h$ attention heads of head size $m$ and hidden layers of width $r$, we state the main theorem as follows, which shows that if the sparse pattern in use satisfies the assumptions below, there exists Diffuser belonging to $\mathcal{D}^{h,m,r}$ that is a universal approximator of continuous sequence-to-sequence function.
\begin{theorem}
\label{thm:main}
Consider any continuous function $f \in \mathcal{F}$, and the class of Diffuser $\mathcal{D}^{h,m,r}$ with sparse attention graph satisfying Assumption \ref{assm:pattern}.
Then, for any $\epsilon >0$ and $1\leq q < \infty$, there exists a function $g \in \mathcal {D}^{h,m,r}$ such that
\begin{equation}
    d_q(f,g) \defeq \left(\int _{\mathbb{D}}\norm{f(\bm{X}) - g(\bm{X})}_q^q d \bm{X} \right)^{1/q} \leq \epsilon.
\end{equation}
\end{theorem}

Compared to other works discussing expressiveness of sparse Transformers \cite{zaheer2020bigbird,yun2020n},
we show Diffuser can achieve contextual mappings using fewer layers based on attention diffusion mechanism.
Intuitively, the improved efficiency can be understood as the expanded attention receptive field through diffusion, which includes more attentions without stacking attention layers.
We then specify a set of conditions on the sparse attention patterns $A$ of the attention graph $G$ in study.
\begin{assumption}
\label{assm:pattern}
Sparsity pattern $A$ satisfies the following:
\begin{enumerate}
  \setlength{\itemsep}{0pt}
    \item \label{assm:pattern-cond1} All tokens attend to themselves, i.e., for all $k \in [n]$, we have $k \in  Ne(k)$.
    \item \label{assm:pattern-cond2} The graph G is connected and has a Hamiltonian path connecting all nodes, i.e., there exists a permutation $\gamma: [n] \to [n]$ such that, for all $i \in [n-1]$, $\gamma(i) \in   Ne({\gamma(i+1)})$.
\end{enumerate}
\end{assumption}

The detailed proof is shown in Appendix \ref{apsec:approximator}, and the key innovation here is that the introduction of attention diffusion allows sequence ID computation to involve all token values within one attention layer.

\subsection{Diffuser as Expander Graphs}
Expander graphs are sparse and robust graphs with strong connectivity, and have several nice properties to improve the expressiveness of Diffuser while keeping the computational efficiency.
In this subsection, we show the sparse attention graph in Diffuser has good expander graph property,
and then highlight three advantages of constructing attention graph as an expander graph, including ensuring sparsity, mixing diffusion rapidly, and approximating full-attention.

We consider the family of $d$-regular graphs $G$ with adjacency matrix $A$, which require all vertices to have the same degree $d$, and we then define the $(\epsilon,d)$-expander:

\begin{definition}
\label{def:expand}
A graph $G$ is a $(d,\epsilon)$-expander if it is d-regular and its adjacency matrix eigenvalues satisfy $|\mu_i|\leq\epsilon d$ for $i\geq2$.
\end{definition}
As the Laplacian eigenvalues of regular graph are given by $\lambda_i=d-\mu_i$, this is equivalent to $|d-\lambda_i|\leq\epsilon d$. We show in Appendix \ref{apsec:expander} the equivalent definitions using expansion ratio and the properties of eigenvalues of $d$-regular graph.
One common random graph model used to build such expander graphs is Erdős–Rényi $\mathcal{G}_{n,p}$ model where each edge is included in the graph with probability $p$, and we consider the variant $\mathcal{G}_{n,m}$ model where $m$ edges are randomly drawn, further constrained to the regularity $d$. These two models are very similar if $p\geq logn/n$ which is satisfied in the long-sequence scenario.
It can be proved that such randomly built $d$-regular is an expander with high probability \cite{friedman2008proof}.
To ensure good expander graph properties, we follow such random models to build the random attention graph which can be thought of as a $(r,\epsilon)$-expander, as discussed in the previous section (additional connections from local and global pattern will not harm expander properties). 

The next theorem shows that such sparse random attention with $(r,\epsilon)$-expander properties can approximate the full-attention complete graph (proved in Appendix \ref{apsec:expander}).
\begin{definition}
\label{def:approximate}
For two graph $G$ and $H$, we say $G$ is an $\epsilon$-approximation to $H$ if $(1+\epsilon)H\succeq G \succeq (1-\epsilon)H$, where $G\succeq H$ means the corresponding Laplacian matrix $L_G-L_H$ is positive-semidefinite.
\end{definition}

\begin{theorem}
\label{thm:approximate}
For every $\epsilon > 0$, there exists $d$ such that for all sufficiently large $n$, there is a $d$-regular graph $G$ which is an $\epsilon$ approximation of the complete graph $K_n$.
\end{theorem}

Further we notice the nice properties from diffusion transformation as low-pass filters can further enhance the expander properties.
The eigenvalues $\tilde{\mu_i}$ of diffusion matrix $\mathcal{A}$ can be computed as $\tilde{\mu_i}=\alpha\sum_{k=0}^{\infty}(1-\alpha)^k\mu_i^k=\frac{\alpha}{1-(1-\alpha)\mu_i}$ in the PPR case,
which amplifies low Laplacian eigenvalues while suppressing high eigenvalues (shown in Appendix \ref{apsec:diff}).

Another essential reason for Diffuser designing expander graphs is that they achieve rapid mixing for random walks and diffusion, which accelerates the information propagation on the attention graph of Diffuser.
\begin{theorem}
\label{thm:mixing}
Given $d$-regular graph with adjacency matrix $A$ and transition matrix $\hat{A}=\frac{1}{d}A$ of random walk, assume the spectral gap $\sigma$ is defined by $\sigma=max(|\mu_2|,|\mu_n|)\triangleq\beta d$. Then,
\begin{equation}
||\hat{A}^tv-u ||_1 \leq \sqrt{n}\beta^t,
\end{equation}
where $ u$ is the stationary distribution and $v$ is an arbitrary initial distribution.
\end{theorem}
The theorem shows that PPR (or general random walk) approaches its limiting probability distribution rapidly on expander graph which has large spectral gap (proved in Appendix \ref{apsec:expander}).
The fast convergence of PPR on expander graph indicates accelerated information propagation in Diffuser.

\section{Experiments}
We evaluate the performance of Diffuser with a rich set of sequence modeling tasks, including language modeling, image modeling and Long-Range Arena (LRA) tasks.
We then analyze expressiveness and efficiency through extensive ablation studies. 

\subsection{Model Implementations}
We implement Diffuser using the graph library DGL
, which offers optimized kernels for sparse matrix operations. 
We first build the graph according to the sparse pattern, then follow the message passing framework defined in DGL by calculating the sparse attention as \textit{message functions}, and attention diffusion as \textit{update and reduce function}.
The remaining components follow the regular Transformer architecture in PyTorch.
The detailed experimental settings, hyperparameters and baseline setup are discussed in Appendix \ref{apsec:exp}.

\begin{figure}[H]
\vspace{-5 pt}
\centering
  \includegraphics[width=0.99\linewidth]{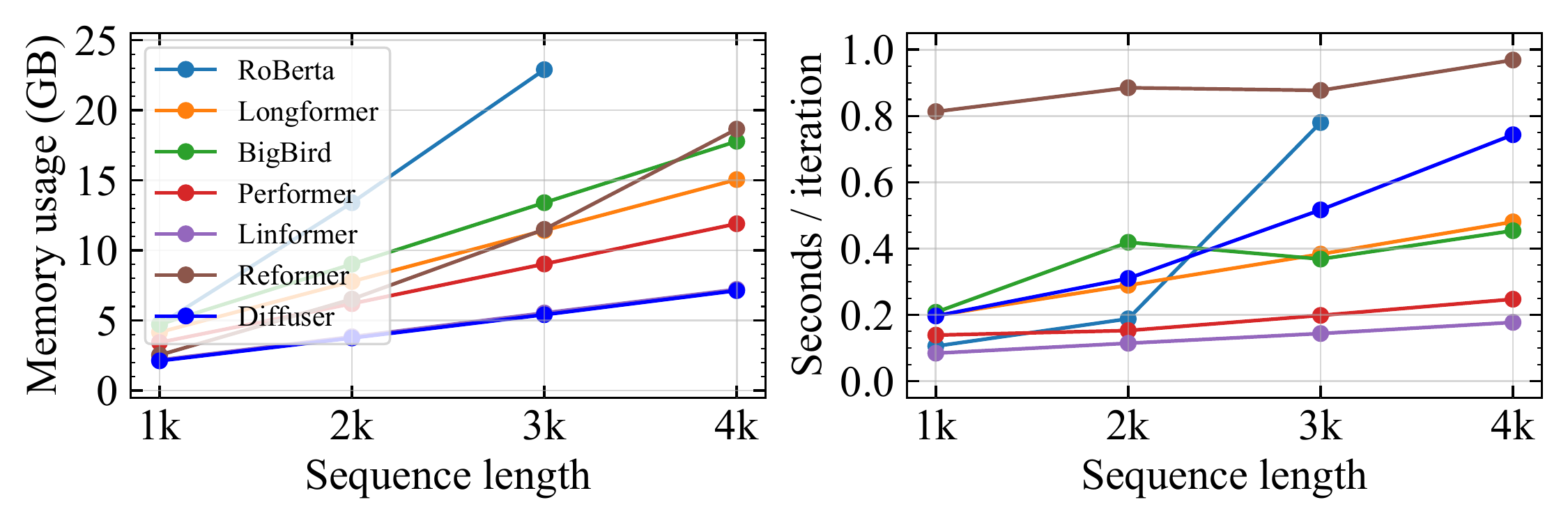}
  \caption{Comparisons of computational efficiency: memory usage and seconds/iteration.  }
  \label{fig:eff}
 \vspace{-5pt}
\end{figure}

\xhdr{Efficiency}
We show the GPU memory usage and runtime comparison in Figure \ref{fig:eff}. 
Compared to benchmarks, Diffuser achieves  1.67$\times$ memory savings compared to the best baseline Performer, with comparable running time. It should be noted that the runtime can be further improved with better diffusion sparse operation support from hardware.

\subsection{Language Modeling}
\xhdr{Pretraining}
We evaluate the model on language tasks following the standard pretraining and finetuning pipeline \cite{roberta2019}. 
Diffuser is pretrained with masked language modeling (MLM) task, which involves predicting a random subset of tokens that have been masked out.
We pretrain the model with three standard datasets  (detailed in Appendix \ref{apsec:exp}) and evaluate the pretraining performance with bits per character (BPC) as in \citet{zaheer2020bigbird}.
The training is conducted with the maximum sequence length of 4,096 and linear warmup from the RoBERTa checkpoint. 
As shown in Table \ref{tab:bpc}, Diffuser achieves lower BPC compared to benchmarks after training for 50K steps. The significant difference between initialization and training 10K steps for Diffuser model indicates the RoBERTa weights are not working well because of the change of the updating rule, and the model is learning to better utilize the attention diffusion.

\begin{table}[h] 
\centering
\small
\begin{tabular}{lc|lc} \toprule
\textbf{Model}     & \textbf{BPC}  & \textbf{Model}     & \textbf{BPC}     \\ \midrule
RoBERTa      &  2.02 & Diffuser-init & 3.52 \\
Longformer   &  1.86 & Diffuser-10K steps & 1.96  \\
BigBird      &  1.82 & Diffuser-50K steps & \textbf{1.68}\\
\bottomrule   
\end{tabular}
\caption{MLM BPC for Diffuser and baselines.}
\vspace{-5 pt}
\label{tab:bpc}
\end{table}

\xhdr{Text classification}
We first evaluate Diffuser on text classification tasks with five datasets. \textbf{Hyperpartisan} \cite{kiesel-etal-2019-semeval} and \textbf{20NewsGroups} \cite{lang1995news} are news datasets with different scales.  \textbf{IMDB} \cite{maas-etal-2011-learning} is a collection of movie reviews for sentiment classification. 
Moreover, we select and propose two new benchmarks with longer documents based on an existing large-scale corpus, Amazon product reviews \cite{he2016modeling}, to conduct long document classification. \textbf{Amazon-512} contains all reviews that are longer than 512 words from the \textit{Electronics} category; \textbf{Amazon-2048} contains 10,000 randomly sampled reviews that are longer than 2,048 words from the \textit{Books} category. 
We randomly split 8/1/1 as train/dev/test sets for both datasets (statistics detailed in Appendix \ref{apsec:exp}).
We finetuned the pretrained Diffuser on each dataset and compare the average F1 score with benchmark models. 
To investigate the influence of attention diffusion, we also apply attention diffusion to Longformer and BigBird models based on their respective sparse patterns (\textit{BigBird\_D} and \textit{Longformer\_D}).
As shown in Table \ref{tab:cls_res}, diffusion-based methods consistently achieve better average score, indicating the importance of attention diffusion.
Among them, Diffuser achieves the best average performance, showing the effect of the proposed sparse pattern on attention diffusion.
Especially, Diffuser outperforms BigBird by 0.4\% and 1.2\% on two long Amazon datasets, which shows its stronger ability to model long sentences. 


\begin{table}[t]
\centering
\resizebox{0.99\columnwidth}{!}{
\begin{tabular}{l|ccccc|c} 
\toprule
             & \textbf{HYP}          & \textbf{20NG}         & \textbf{IMDB}         & \textbf{A-512}        &\textbf{A-2048}       & \textbf{Avg.}  \\ \midrule 

95pt. & 2,030        & 1,229         & 771      & 1,696  &  5,216  & -  \\ \midrule
BERT    & 85.7        & 85.3        & 91.3        & 59.2        & 50.3        & 74.36  \\
RoBERTa & 87.4        & 85.7        & 95.3        & 65.0        & 57.9        & 78.26  \\
BigBird      & 92.2        & 82.3        & 95.2        & 67.4        & \underline{63.6}       & 80.14  \\
Longformer    & \underline{93.8}        & 86.3        & \textbf{95.7}        & 67.3        & 61.2        & 80.86  \\
\midrule
BigBird\_D     &93.1         & 84.5              & 95.0             &  \textbf{68.2}   &63.4&  80.84\\
Longformer\_D  &93.5         &  \textbf{87.3}    & \underline{95.4} & 67.0             &62.5            &  \underline{81.24}\\
Diffuser      &\textbf{94.4}& \underline{86.8}  & 95.2             & \underline{67.8} & \textbf{64.8}  &  \textbf{81.80}                                  \\
\bottomrule
\end{tabular}
}
\vspace{-1mm}
\caption{Text classification results on five datasets: Hyperpartisan (HYP), 20NewsGroups (20NG), IMDB, Amazon-512 (A-512) and Amazon-2048 (A-2048). \texttt{95pt.} indicates 95th percentile of token number. We report average F1 scores (Avg.). We underscore the best among baselines, and bold the best overall models.}
\label{tab:cls_res}
\vspace{-20 pt}
\end{table}

\xhdr{Question answering}
We choose two benchmarks for question answering: WikiHop \cite{welbl2018wikihop} and TriviaQA \cite{joshi-etal-2017-triviaqa}. WikiHop is a dataset collected based on Wikipedia articles for multi-hop question answering across documents. TriviaQA is a large-scale dataset of question-answer-evidence pairs for reading comprehension. Both datasets are in a reasonable scale and length, as in Table \ref{tab:qa_res}. 
We follow \citet{beltagy2020longformer} and concatenate the question, answer, and candidates into one input sequence with special tokens along the context.
Task specific projection layers are then adopted to classify the correct answers for WikiHop and predict the answer span for TriviaQA.
From Table \ref{tab:qa_res}, we see that Diffuser achieves the best results for TriviaQA and has comparable performance for WikiHop.

\begin{table}[h]
\centering
\small
\resizebox{0.49\textwidth}{!}
{
\begin{tabular}{lrrr}
\toprule
    \textbf{Stats.}   & \textbf{WikiHop}  & \textbf{TriviaQA}  \\ 
    \midrule
    Avg  & 1564.48            & 11641.95               &   \\
    95pt.  & 3672            & 32,158               &   \\ \midrule
    Train  & 43,738            & 61,888              &   \\
    Val  &5,129            & 7,993              &   \\
    Test  & -            & 7,701              &   \\
  \bottomrule
\end{tabular}

\quad

\begin{tabular}{lccc}
\toprule
   \textbf{Model}   & \textbf{WikiHop}  & \multicolumn{2}{c}{\textbf{TriviaQA}}    \\ 
   \textbf{Metric}                 & \textbf{Acc}      & \textbf{F1} & \textbf{EM}   \\
    \midrule
    RoBERTa          & 71.82  & 74.02 & 66.87            \\
    Longformer       & 75.30  & 74.82 & 67.24        \\
    BigBird          & 74.54  & 73.16 & 68.26        \\
    \midrule           
    Diffuser          & \textbf{75.80}  & \textbf{75.84} & \textbf{70.20}  \\
    \bottomrule
\end{tabular}
}
\caption{Comparison of WikiHop and TriviaQA, and model performances. We report Accuracy for WikiHop, and F1, EM score for TriviaQA.} 
\label{tab:qa_res}
\vspace{-10 pt}
\end{table}

\subsection{Image Generative Modeling}
We then evaluate the performance of Diffuser on image density modeling task with CIFAR-10 and ImageNet-64. 
The sequence lengths are 3,072 and 12,288, respectively.
We follow the setting of \cite{child2019generating} and adopt an 8-layer model with 512 hidden dimensions which is trained until the validation errors stop decreasing. 
As shown in Table \ref{tab:img}, Diffuser achieves lower bits per dimension (BPD) on CIFAR-10 datasets and converges to similar BPD on ImageNet64 dataset, demonstrating the effectiveness of the model in the image domain. Similar results are obtained with different layers and hidden dimensions.
\begin{table}[h] 
\centering
\resizebox{0.8\columnwidth}{!}{
\begin{tabular}{lc|lc} \toprule
\textbf{CIFAR-10}     & \textbf{BPD}  & \textbf{ImageNet-64}     & \textbf{BPD}     \\ \midrule
PixcelCNN      &     3.03 & PixcelCNN & 3.57 \\
PixcelCNN+     &      2.92 & Parallel Multiscale & 3.70  \\
PixelSNAIL     &       2.85 & SPN & 3.52
\\ Sparse Trans. &  2.80 &  Sparse Trans. & \textbf{3.44}\\
\midrule
Diffuser       &  \textbf{2.78} & Diffuser  & \textbf{3.44}\\
\bottomrule
\end{tabular}
}
\caption{Bits per Dimension (Bits/Dim) on CIFAR-10 and ImageNet-64. We list baseline details in Appendix \ref{apsec:exp}.}
\vspace{-5 pt}
\label{tab:img}
\end{table}

\vspace{-10 pt}
\subsection{Long-Context Sequence Modeling} 
Long Range Arena (LRA) \cite{tay2021long} is a unified benchmark for evaluating efficient Transformer models with five multi-class classification tasks from different domains, including ListOps, byte-level text classification, byte-level document retrival, image classification, and image-based path finder. 
All the tasks are multi-class classification with input sequences of different lengths.
As shown in Table \ref{tab:lra_res}, Diffuser achieves the best results on ListOps (2K), Retrieval (4K), and Image (1K), improving average accuracy by 2.30\% compared to the best benchmark BigBird.

\begin{table}[h]
\centering
\resizebox{1\columnwidth}{!}{
\begin{tabular}{l|ccccc|c}
  \toprule
    \textbf{Models} &  \textbf{ListOps} & \textbf{Text}  & \textbf{Retrieval} & \textbf{Image} & \textbf{Pathfinder} & \textbf{Avg} \\ 
\midrule
Transformer & 36.37 & 64.27& 57.46 & 42.44 & 71.40 &  54.39\\
\midrule
    Local Attention                             & 15.82             & 52.98                 & 53.39                                                & 41.46             & 66.63                 & 46.06 \\
    Linear Trans.          & 16.13             & \textbf{65.90} & 53.09  & 42.34             & 75.30                 & 50.55 \\
    Reformer                   & \underline{37.27} & 56.10                 & 53.40  & 38.07             & 68.50                 & 50.67 \\
    Sparse Trans.    & 17.07             & 63.58                 & 59.59  & \underline{44.24} & 71.71                 & 51.24 \\
    Sinkhorn Trans.            & 33.67             & 61.20                 & 53.83   & 41.23             & 67.45                 & 51.48 \\
    Linformer                   & 35.70             & 53.94                 & 52.27   & 38.56             & 76.34                 & 51.36 \\
    Performer                & 18.01             & \underline{65.40}        & 53.82   & 42.77             & \textbf{77.05}     & 51.41 \\
    Synthesizer              & 36.99             & 61.68                 & 54.67        & 41.61             & 69.45                 & 52.88 \\
    Combiner &36.65          &64.99   &\underline{59.81}         &41.67         &71.52   & 54.93        \\ 
    Longformer      & 35.63             & 62.85                 & 56.89        & 42.22             & 69.71                 & 53.46 \\

    BigBird             & 36.05             & 64.02                 & 59.29    & 40.83             & 74.87                 & \underline{55.01} \\

\midrule
    Diffuser                             & \textbf{37.52}    &   65.20               & \textbf{61.28}   
                                                & \textbf{45.20}     & \underline{76.58}       & \textbf{57.31} \\
\bottomrule
\end{tabular}
}
\vspace{-1mm}
\caption{Classification accuracy on LRA datasets with three best performing benchmarks on average. Underline values are best among baselines, while bold are the best.
}
\label{tab:lra_res}
\vspace{-4mm}
\end{table}

\subsection{Ablation Studies}
\begin{figure}[h]
\centering
  \includegraphics[width=0.99\linewidth]{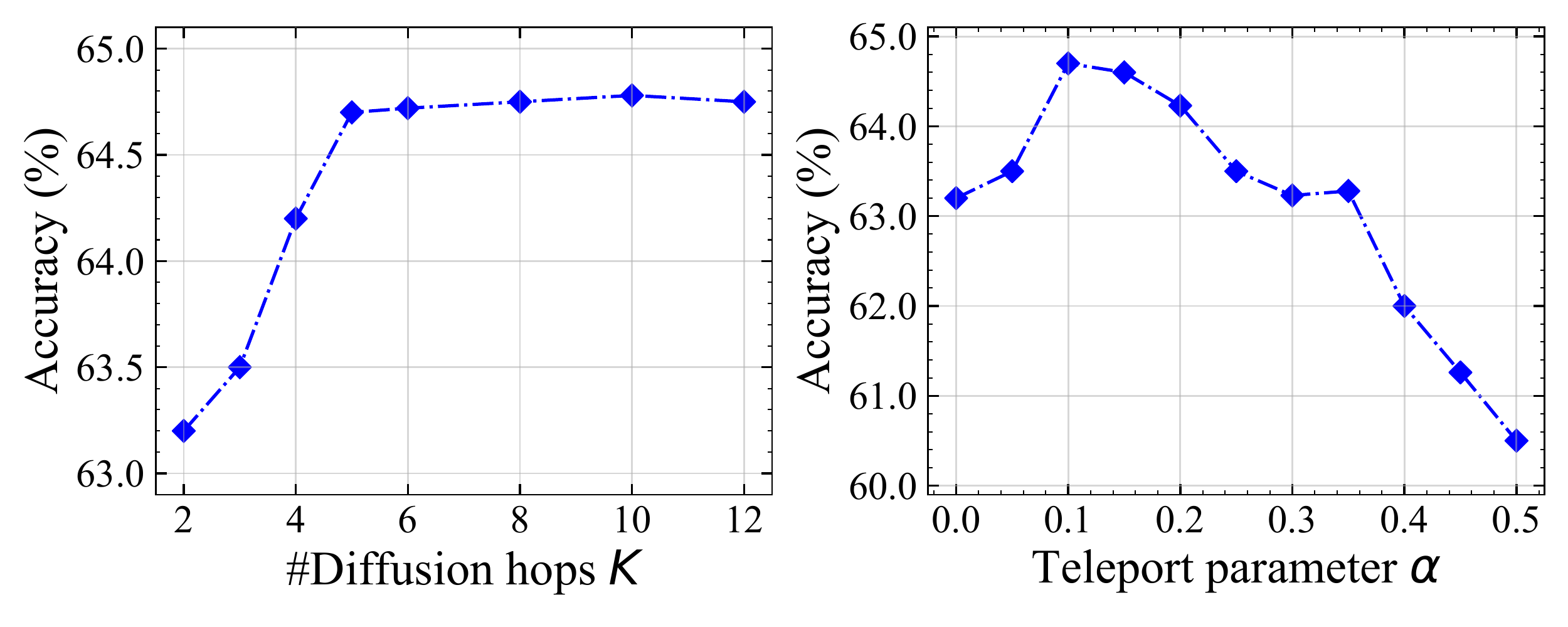}
  \vspace{-5 pt}
  \caption{The influence of diffusion parameters. }
  \label{fig:diff}
  \vspace{-15 pt}
\end{figure}

\begin{figure}[h]
\centering
  \includegraphics[width=0.99\linewidth]{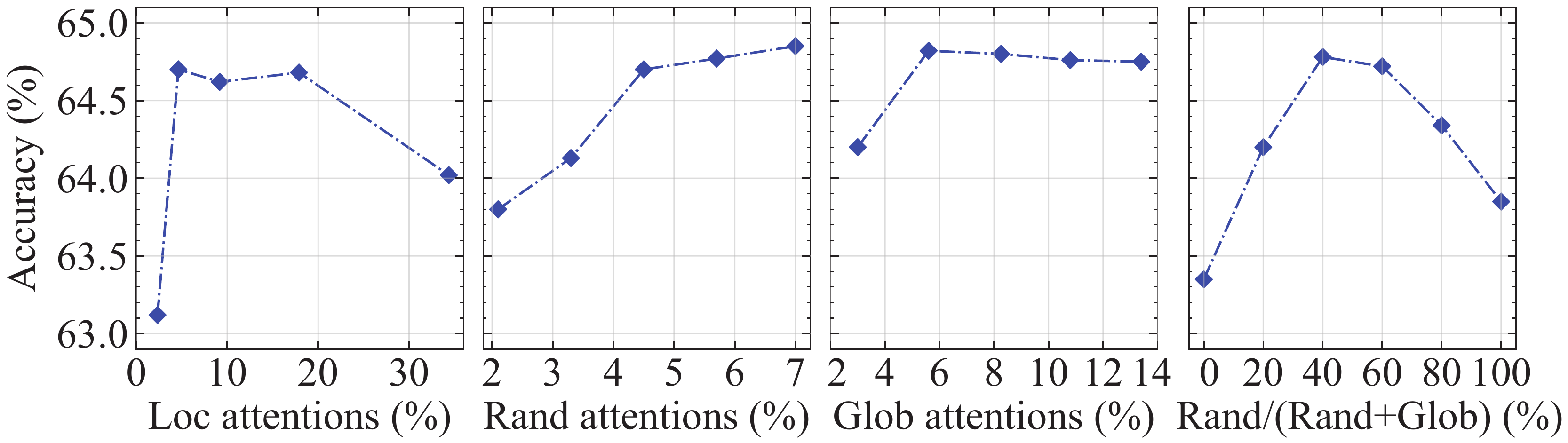}
  \vspace{-5 pt}
  \caption{The influence of different types of attentions. }
  \label{fig:atts}
 \vspace{-10 pt} 
\end{figure}

We first study the influence of different mechanisms used in Diffuser by ablating the corresponding components. Table \ref{tab:abl} shows that the diffusion (\#4)  and local patterns (\#1)  have the biggest influence on the performance while random (\#2)  and global attentions (\#3)  result in similar performance drop.
We also notice that changing the random attention into uniform distribution (\#5) does not substantially affect the performance as the expansion properties are retained (shown in Appendix \ref{apsec:sparsity}).
We then investigate the effect of the diffusion parameter using A-2048 datasets as shown in Figure \ref{fig:diff}. We observe significant improvement in performance as $K$ increases, and the saturated performance under $K\geq 5$ indicates the convergence to the stationary distribution. 
We also observe the performance is significantly influenced by the teleport parameter $\alpha$, and we choose $K=5$ and $\alpha=0.1$ in practice.
We also show the influence of different types of attentions in Figure \ref{fig:atts}. The performance improvements slow down as we increase the number of attentions for all three types of attentions, and we choose the number of attentions considering the balance between expressiveness and efficiency.
We also observe that there exists an optimal ratio to combine the random and global attentions and  improve performance upon  random-attentions-only or  global-attentions-only scenarios.
More ablation studies (e.g., input robustness analysis) are shown in Appendix \ref{apsec:ablations}.

\begin{table}[t]
    \centering
\begin{center}
\resizebox{0.45\textwidth}{!}
{
\begin{tabular}{l|ccccc|c}
\toprule
    \textbf{Model} &  \textbf{ListOps} & \textbf{Text}  & \textbf{Retrieval} & \textbf{Image} & \textbf{Pathfinder} & \textbf{Avg} \\ 
\midrule
    \#0 Diffuser    & \textbf{37.52} & \underline{65.20} &\underline{61.28} & \textbf{45.20} & \textbf{76.58} & \textbf{57.31} \\
    \midrule
    \#1 w/o loc.        & 35.28 & 58.60 & 58.05 & 38.07 & 73.25 & 52.65 \\
    \#2 w/o rand.  & 36.38 & 64.89 & 60.60 & 43.48 & 73.36 & 55.74 \\
    \#3 w/o glob.  & 36.52 & 63.53 & 61.07 & 42.35 & 72.47 & 55.19 \\    
    \#4 w/o diff.    & 33.48 & 59.73 & 58.25 & 39.28 & 71.08 &52.36 \\
    \#5 w uni. & \underline{37.39} & \textbf{65.25} & \textbf{61.42} & \underline{44.98} & \underline{76.30} & \underline{57.07} \\
\bottomrule
\end{tabular}
} 
\end{center}
\setlength{\abovecaptionskip}{-0.8mm}
\caption{Ablation studies of each component in Diffuser.}
\label{tab:abl}
\vspace{-10 pt}
\end{table}



\section{Conclusion}

In this work, we proposed Diffuser, an efficient Transformer for long sequence modeling that applies multi-hop attention diffusion. We theoretically showed that Diffuser is a more efficient universal approximator for sequence modeling, with better expander properties from the graph spectral perspective. Experimentally, we showed that Diffuser achieves superior performance in language modeling, image modeling, and other long sequence modeling tasks.

\bibliography{aaai22}

\title{Appendix}

\lstset{%
	basicstyle={\footnotesize\ttfamily},
	numbers=left,numberstyle=\footnotesize,xleftmargin=2em,
	aboveskip=0pt,belowskip=0pt,%
	showstringspaces=false,tabsize=2,breaklines=true}
\floatstyle{ruled}
\newfloat{listing}{tb}{lst}{}
\floatname{listing}{Listing}

\pdfinfo{
/Title ()
/Author ()
/TemplateVersion (2022.1)
}

\setcounter{secnumdepth}{1} 

\renewcommand{\thesection}{\Alph{section}}
\renewcommand{\thesubsection}{\thesection.\alph{subsection}}

\onecolumn
\section{Attention Diffusion Analysis}
\label{apsec:diff}
We adopt the following power iteration equation for attention diffusion and approximate $\mathcal{A}V$.
\begin{equation}
\label{eq:propagate}
\begin{split}
    Z_{(0)}&=V=XW_V,\\
    Z_{(k+1)}&=(1-\alpha)AZ_{(k)}+\alpha V\nonumber.
\end{split}
\end{equation}

\begin{proposition}\label{prop:approxGAT_appendix}
    $\lim_{K\rightarrow\infty}Z_{(K)} = \mathcal{A}V$ 
\end{proposition}

\xhdr{Proof} Let $K > 0$ be the total number of iterations (i.e., number of hops in graph attention diffusion).
After $K$-th iteration, we can get
\begin{equation}
    Z_{(K)} = \left((1 - \alpha)^{K}A^{K} + \alpha\sum_{i=0}^{K-1}(1-\alpha)^{i}A^{i}\right)V.
    \nonumber
\end{equation}
The term $(1 - \alpha)^{K}A^{K}$ converges to 0 as $\alpha \in (0,1]$ and $A^{K}_{i,j} \in (0,1]$ when $K \rightarrow \infty$, and thus $\lim_{K\rightarrow\infty}Z^{(K)} = \left(\sum_{i=0}^{\infty}\alpha(1-\alpha)^{i}A^{i}\right)V$ = $\mathcal{A}V$.

\subsection{Influence of Diffusion on Graph Spectral Properties}
Diffusion with either PPR or heat kernel acts as low-pass filters, which amplifies small eigenvalues which represents the large-scale structure in the graph and decreases large eigenvalues corresponding to noise in the graph.
We define the normalized graph Laplacian for $A$ and $\mathcal{A}$ as $L = I-A$ and $\mathcal{L} = I-\mathcal{A}$ (because attention matrix has degree 1 for each node), with corresponding eigenvalues $\lambda$ and $\tilde{\lambda}$, then we have: 
\begin{proposition}
$L$ and $\mathcal{L}$ have the same set of eigenvectors. If $\lambda_i$ and  $\tilde{\lambda}_i$ are the eigenvalues corresponding to the $i$-th eigenvector,  we have 
\begin{equation}
    \tilde{\lambda}_i = 1- \frac{\alpha}{1-(1-\alpha)(1-\lambda_i)}.
\end{equation}
\end{proposition}
\xhdr{Proof}
The eigenvalues $\tilde{\mu_i}$ of diffusion matrix $\mathcal{A}$ can be computed as $\tilde{\mu_i}=\alpha\sum_{k=0}^{\infty}(1-\alpha)^k\mu_i^k=\frac{\alpha}{1-(1-\alpha)\mu_i}$.
Then combining $L = I-A$ with $\mathcal{L} = I-\mathcal{A}$, we can obtain the Laplacian eigenvalues. Because the eigenvalue of the normalized adjacency matrix (attention matrix) is in the range  $\mu_i\in[-1,1]$, we have $\lambda_i\in[0,2]$.  

\begin{figure}[b]
\centering
\includegraphics[width=0.6\linewidth]{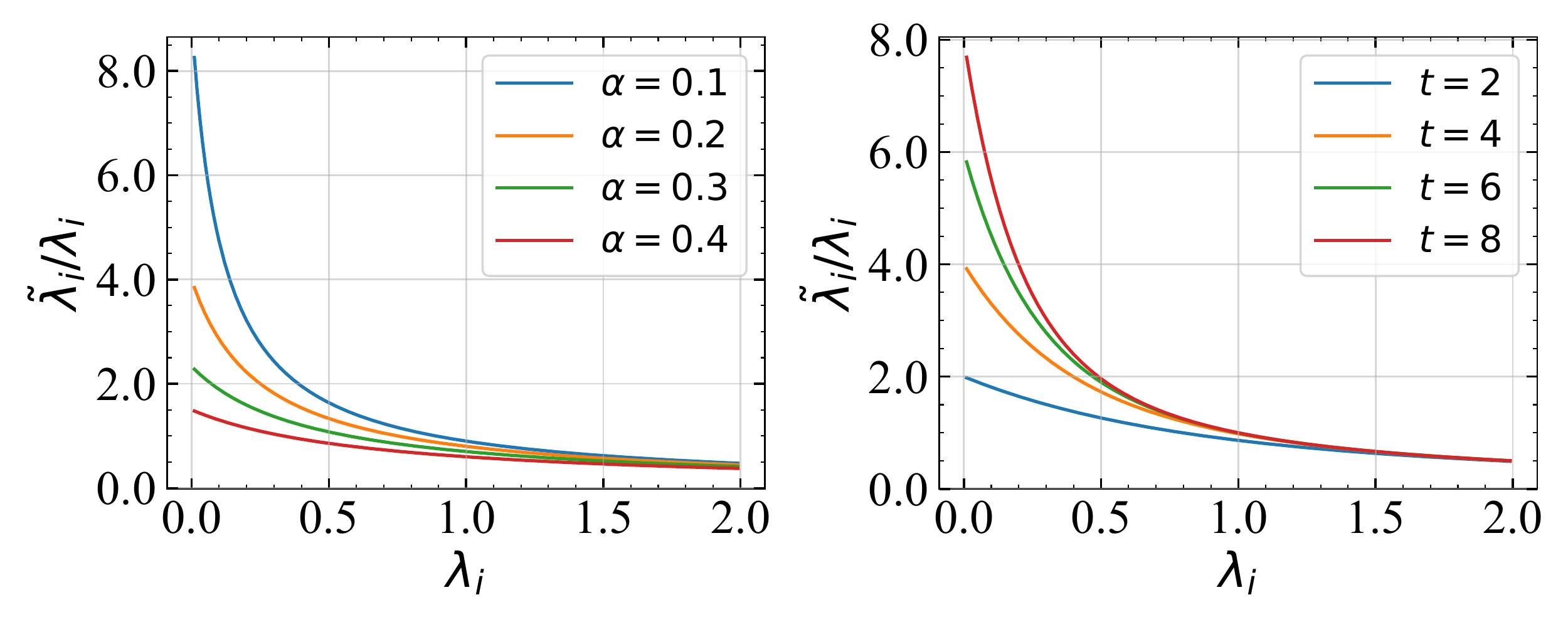}
\caption{Influence of diffusion with PPR kernel (left) and heat kernel (right) on the graph spectral properties.  }
\label{apfig:spectral}
\vspace{-5pt}
\end{figure}

Similarly, we have $ \tilde{\lambda}_i = 1- e^{-\lambda_i t}$ if using heat kernel with temperature $t$ instead. We then visualize the effect of diffusion on the graph spectral properties. As shown in Figure \ref{apfig:spectral}

\section{Sparse Pattern Analysis}
\subsection{Sparsity Analysis}
We analyze attention pattern sparsity of different models, and show the contributions of each type of attention in Table \ref{aptab:pattern}. We see Diffuser achieves the most sparse attention pattern for all input lengths. For BigBird and Diffuser, local window attentions contribute the most with sequence length 1,024 and 2,048, and all three types of attentions contribute almost equally for lengths of 4,096 and 8,192.
\label{apsec:sparsity}
\begin{table}[H]
\centering
\begin{tabular}{ccccll}
\toprule
\textbf{Length}                & \textbf{Longformer} & \textbf{BigBird}          & \textbf{Diffuser}       &  &  \\ \midrule
\multirow{2}{*}{1024} & 62.5       & 55.7             & 24             &  &  \\
                      & (62.5/-/-) & (26.2/16.4/13.0) & (18.0/4.2/1.9) &  &  \\ \midrule
\multirow{2}{*}{2048} & 34.4       & 32.5             & 15.5           &  &  \\
                      & (34.4/-/-) & (13.6/10.4/8.5)  & (9.2/.4.2/2.1) &  &  \\ \midrule
\multirow{2}{*}{4096} & 18         & 16.9             & 11.2           &  &  \\
                      & (18.0/-/-) & (6.9/5.7/4.3)    & (4.6/4.3/2.2)  &  &  \\ \midrule
\multirow{2}{*}{8192} & 9.2        & 8.7              & 5.7            &  &  \\
                      & (9.2/-/-)  & (3.5/2.3/3.0)    & (2.3/2.2/1.1)  &  &  \\ \bottomrule
\end{tabular}
\caption{The percentage of attentions with respect to full-attention (non-zero entries in the sparse pattern) with different input lengths. For each entry, the first row is the total attentions, the second row is (local/ global/ random attentions)}.
\label{aptab:pattern}
\end{table}

\subsection{Spectral Analysis}
To better understand the characteristics of different attention patterns, we analyze and compare the corresponding spectral properties.
We compute the spectrum (eigenvalue v.s. the eigenvalue index) of normalized Laplacian of the graph defined by the attention pattern, with maximum sequence length 2,048, and  therefore the graph has 2,048 nodes and 2,048 eigenvectors. 
First we observe some common properties: all graphs have smallest eigenvalues 0, which indicates all graphs are connected. All graphs eigenvalues are within the range $[0,2]$ which is the property of the normalized graph Laplacian.

\begin{figure}[h]
\centering
\includegraphics[width=0.9\linewidth]{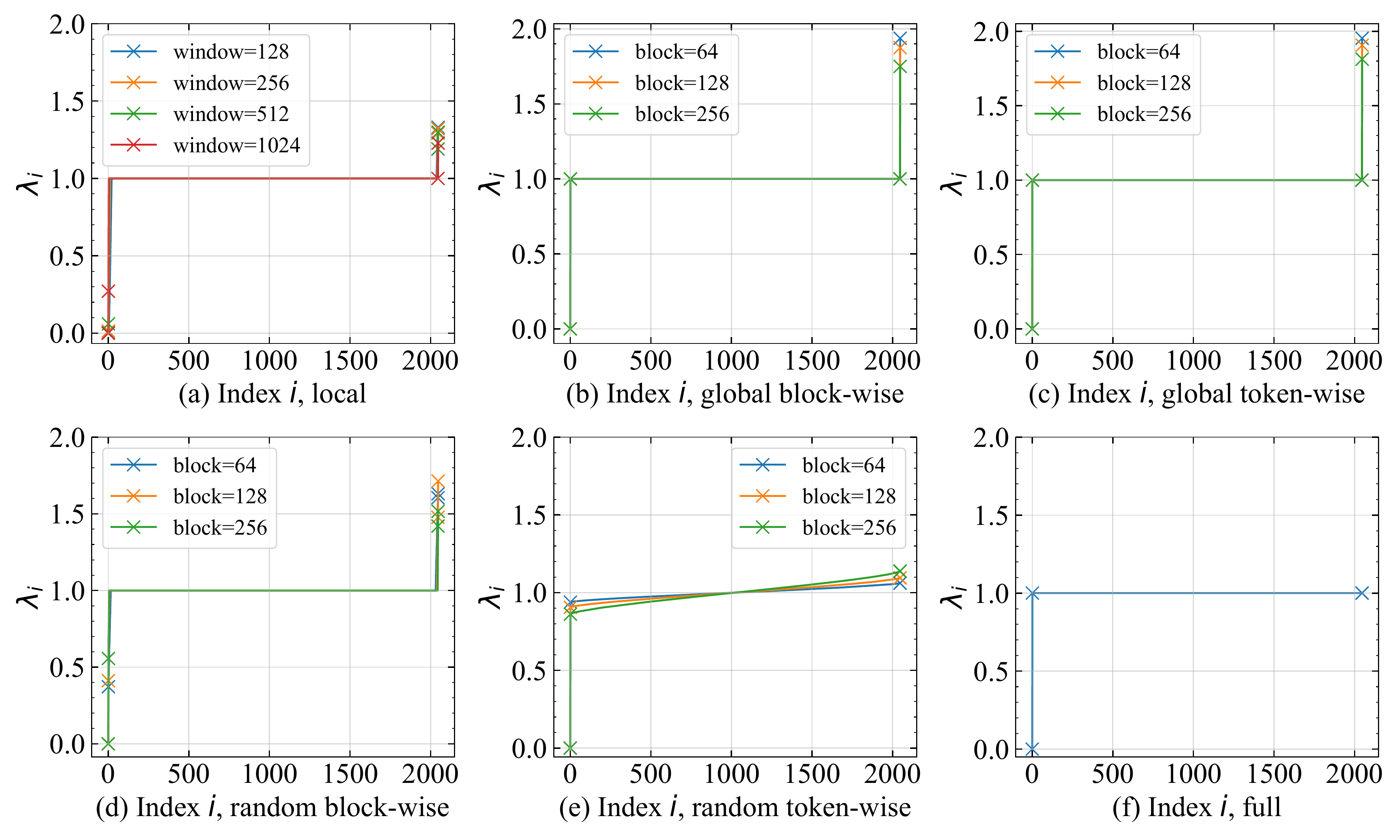}
\caption{The spectra of different types of attentions. 
The first-smallest, second-smallest second-largest, and first-largest eigenvalues $\lambda_1$, $\lambda_2$, $\lambda_{n-1}$, $\lambda_n$ are marked.
(a) The spectra of local window attentions with different window sizes. 
(b)(c) The spectra of global attentions constructed by block-wise selection (used in BigBird) and token-wise selection (used in Diffuser) with different block sizes (the word "block size" is used in token-wise selection to match the number of attentions in the block-wise selection case). 
(d)(e) The spectra of random attentions constructed by block-wise selection (used in BigBird) and token-wise selection (used in Diffuser) with different block sizes. 
(f) The spectrum of full-attention (complete graph).
}
\label{apfig:com}
\vspace{-5pt}
\end{figure}

We compare the spectra of local window attention graph, global attention (block-wise and token-wise) graph, random attention (block-wise and token-wise) graph, and full-attention (complete graph), as in Figure \ref{apfig:com}. 
Cheeger's inequality implies the expansion ratio is bounded by $\lambda_2$ on both sides.
Therefore, we observe that (1) global and random attentions achieve better expansion ratio than local window attentions. 
(2) Token-wise selection can improve the graph expander properties compared with block-wise selection. 
(3) The spectra of local window and random attentions are more similar to full-attention for large eigenvalues, compare with global attentions. 
Therefore, we design the sparsity pattern of Diffuser leveraging advantages of all three types of attentions.

\begin{figure}[h]
\centering
\includegraphics[width=0.9\linewidth]{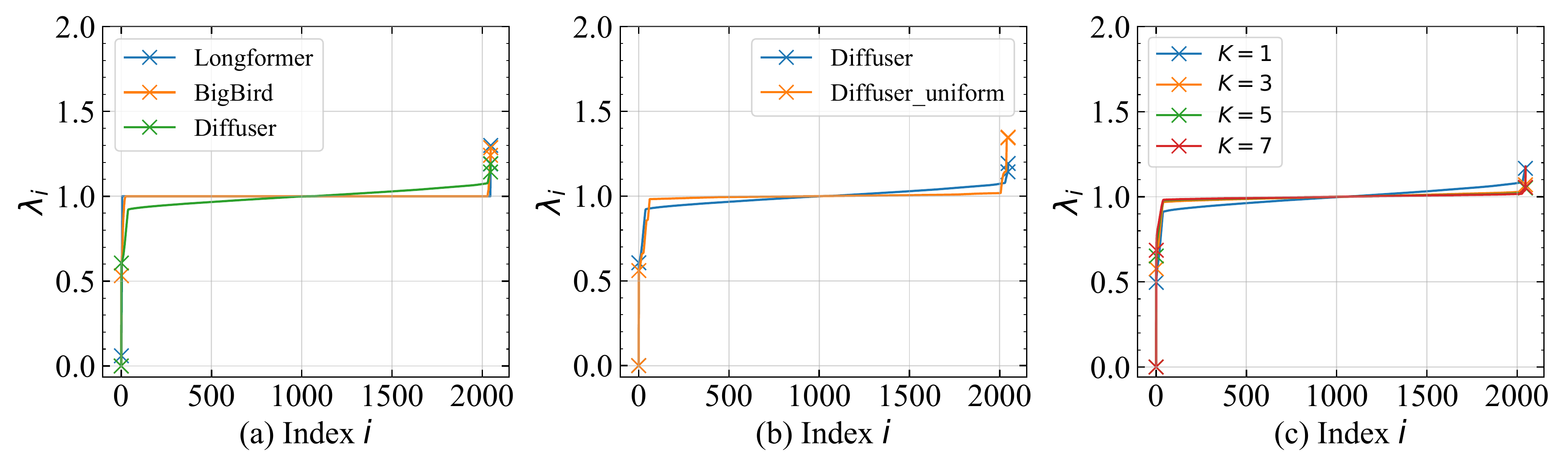}
\caption{The spectra of different models and model variants. 
The first-smallest, second-smallest second-largest, and first-largest eigenvalues $\lambda_1$, $\lambda_2$, $\lambda_{n-1}$, $\lambda_n$ are marked (except for (c) where $\lambda_1$ is omitted for visibility).
(a) The spectra of Longformer, BigBird, Diffuser sparse patterns.
(b) The spectra of Diffuser with original randomly-selected random attentions, and \lq\lq random" attentions constructed by uniform separations.
(c) The spectra of Diffuser with different diffusion steps.
}
\label{apfig:diffcom}
\vspace{-5pt}
\end{figure}

Then we compare the spectra of different models and model variants. As shown in Figure \ref{apfig:diffcom}, (a) Diffuser has larger spectral gap and better expansion properties compared with BigBird and Longformer. (b) Randomness can improve the model expansion property. (c) The expansion ratio will increase with more diffusion steps, and will saturate when approaching the stationary states when $K$ is large.

\section{Diffuser as Universal Approximators}
\label{apsec:approximator}
We consider the input $\mX=[x_1,...,x_n]\in \mathbb{R}^{d\times n}$, 
let $\mc F$ be the class of continuous functions $f: \sD \to \reals^{d \times n}$ defined on any compact domain $\sD \subset \reals^{d \times n}$, where continuity is defined with respect to the entry-wise $\ell_p$ norm ($1\leq p < \infty$). 
We then formulate the self-attention mechanism for each token at layer $l$ as 
\begin{align}
    Attn^l(\mX) &= \mX + \mW_O 
    \begin{bmatrix}
    head^{1,l}(\mX)\\
    \vdots\\
    head^{h,l}(\mX)
    \end{bmatrix}\!,
    \\
    out^l(\mX) &= Attn^l(\mX) + \mW_2 \cdot \text{ReLu} (\mW_1 \cdot Attn^l(\mX)).
\end{align}
The token-wise form to calculate each token $i$ at $m$-th head is 
\begin{align}
    ~~head^{m,l}(\mX)_i = \mW_V^i \mX_{Ne(i)} \cdot \rho [(\mW_K^m \mX_{Ne(i)})^T \mW_Q^m \mX_i].
\end{align}
According to the attention diffusion, the updating rule for each head in the sparse attention can be rewritten as 
\begin{align}
\label{apeq:attdiff}
    \mZ_i^{(0)}=\mW_V^m \mX_{i},  \ \ \  \mZ_i^{(k)}&= (1-\alpha)\mZ_{Ne(i)}^{(k-1)} \cdot \rho [(\mW_K^m \mX_{Ne(i)})^T \mW_Q^m \mX_i] + \alpha \mZ_i^{(0)}.
\end{align}
Then the final output of the head $m$ equals to $head^{m,l}(\mX)_i = \mZ_i^{(K)}$ after $K$ diffusion steps.

\subsection {Proof}
We follow the three-step procedure to show Diffuser's power for universal approximation in the sequence modeling, and the key innovation is in Step 2 where we apply the attention diffusion mechanism.
\subsubsection{Step 1: Approximating continuous function with piece-wise constant function}
The function to be approximated is the continuous function $f\in \mathcal{F}$, which is a mapping from the bounded domain $\mathbb{D}\subset{[0,1)^{d\times n}}$ to the output domain $\mathbb{R}^{d\times n}$.
We approximate $f$ with a piece-wise constant function $\overline{f}$ by partitioning the $[0,1)$ into grid $\mathbb{G}_\delta$ of granularity $\delta$ , $\mathbb{G}_\delta\defeq\{0,\delta,...,1-\delta\}$, such that $d(f,\overline{f})\leq \epsilon/2$ by choosing small enough $\delta$.
This step is also commonly adopted to prove the universal theorem for other general neural networks.
Specifically, we define the piece-wise continuous function family  $\overline {\mc F}(\delta)$,
\begin{equation*}
    \overline{\mc F}(\delta) \defeq \left \{ \mZ \mapsto \sum_{\mG \in \sG_\delta} \mA_\mG \indic{\mZ \in \mG+[0,\delta)^{d\times n}}  \mid \mZ \in \sD, \mA_\mG \in \R^{d\times n}\right \}.
\end{equation*}
Then we have the following Lemma from Lemma 8 of \citet{yun2019transformers}
\begin{lemma}
\label{splem:f_fbar}
For any $f \in \mc F$ and $\epsilon>0$, there exists a small enough $\delta > 0$ such that there exists $\overline f \in \overline{\mc F}(\delta)$ such that $d_q(f,\overline f) \leq \epsilon/2$.
\end{lemma}

\subsubsection{Step 2: Construction of context mapping using attention diffusion}
To accommodate the piece-wise constant function $\overline{f}$, we then slightly modify Diffuser layer by replacing ReLU nonlinearities  and softmax attention score with piece-wise linear activation $\phi\in\Phi$ and hardmax operators $\sigma()$, respectively, resulting in the modified Diffuser architecture $\overline{\mathcal{D}}^{h,m,r}$.
Constructing the contextual mapping is the key challenge of the proof, which maps tokens in different sequences to different values.
We first add appropriate positional encodings and quantize the inputs with feed-forward layers, and then construct a contextual mapping from quantized inputs to unique sequence ids, using both sparse attention and attention diffusion.
Sequence ids are then mapped to the desired outputs through stacked feed-forward layers.
Formally, we will prove the following Lemma in this step:
\begin{lemma}
\label{splem:main}
For any $\overline f \in \overline{\mc F}(\delta)$, there exists $\overline g \in \overline{\mc {D}}^{2,1,1}$ such that $\overline{f}(\mX) = \overline{g}(\mX)$ for all $\mX \in \sD$.
\end{lemma}

The remaining of this step is to prove this lemma, where we utilize the attention diffusion mechanism for contextual mappings.

We first add appropriate position embedding $\mE$ to make tokens follow specific order using certain permutations. Recall from the Assumption \ref{assm:pattern} that there exits a permutation $\gamma:[n]\to [n]$ such that for all $i\in [n-1]$, $\gamma(i)$ is one of the tokens that $\gamma(i+1)$ attends to. We then choose the columns of positional embedding $E$ as follows:
\begin{equation}
    \mE_{\gamma(1)} = (n-1)\ones_n, \text{ and } \mE_{\gamma(i)} = (i-2)\ones_n, \text{ for } i \in [2:n].
\end{equation}
Therefore, token $\gamma(1)$-th column of $\mX+\mE$ will be in the range $[n-1,n)^d$, and similarly $\mX_{\gamma(i)} + \mE_{\gamma(i)} \in [i-2,i-1)^d$ for $i \in [2:n]$. 
This will allow entries of different tokens in the sequences lie in disjoint intervals $[i-2,i-1)$ for $j\in[0:n-1]$.

Next, we quantize the input $\mX+\mE$ using a group of modified feedforward layers to an element $\mZ$ on the grid $\mathbb{Z}_{\delta}\defeq\{0,\delta,...,n-\delta\}^{d\times n}$ (Lemma 5 in \citet{yun2019transformers}).

 \begin{wrapfigure}{h}{0.3\textwidth}
    \centering
    \vspace{-0.2in}\includegraphics[width=1.0\linewidth]{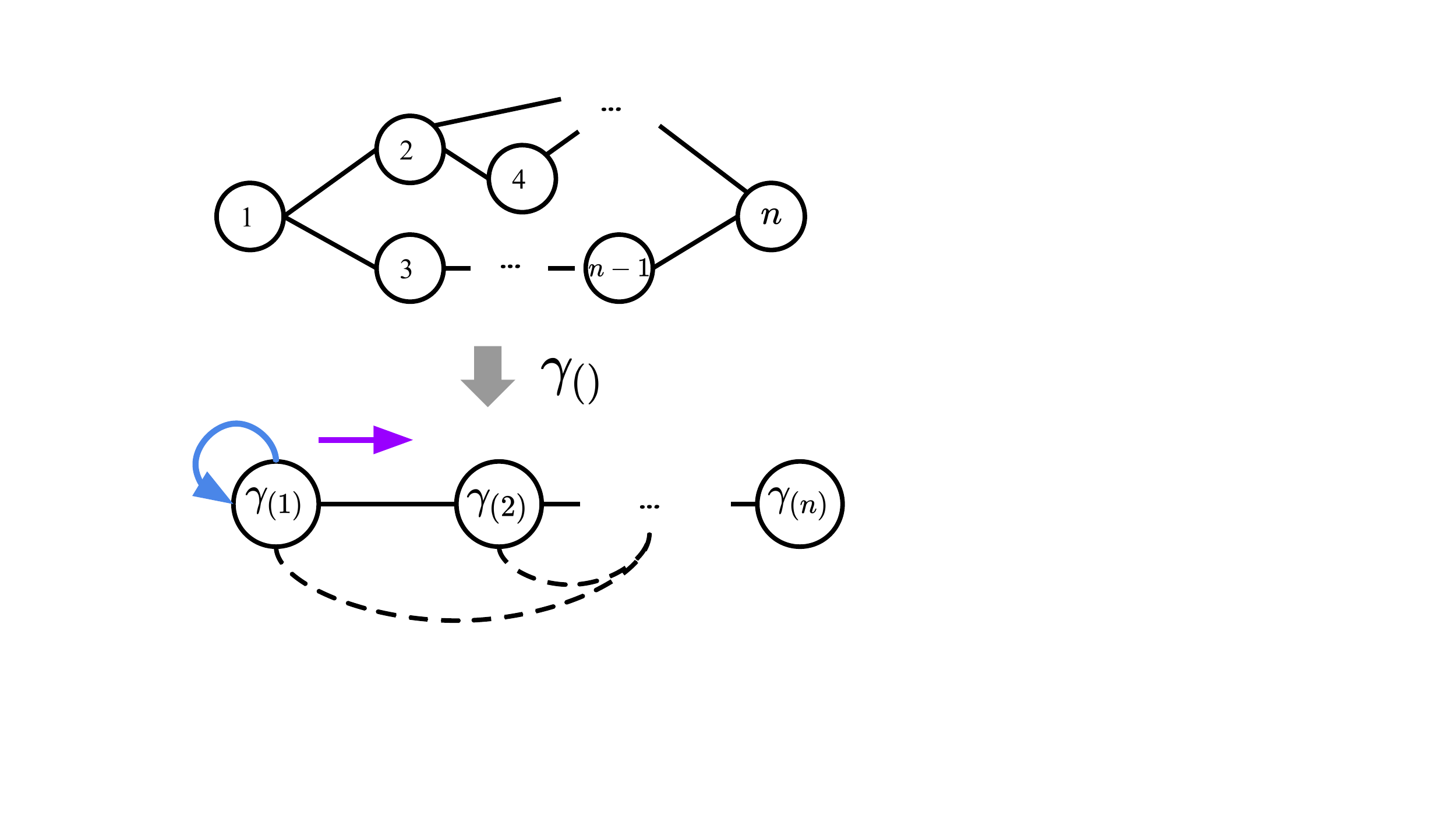}
    \vspace{-0.02in}\caption{The order of the graph nodes before and after permutation $\gamma$.}
    \label{apfig:gamma}
    \vspace{-5mm}
\end{wrapfigure}

Based on the quantized inputs,we are then going to prove the modified Diffuser can realize contextual mappings from the quantized input $\mZ$ to a unique context id associated with the sequence, with the following Lemma:
\begin{lemma}
\label{aplm:context}
Suppose the modified Diffuser layers ($h=2, m=1$) has sparse patterns satisfying Assumption \ref{assm:pattern}  and employs the hardmax $\sigma_H$ operator.
Then there exists function $g_c: \mathbb{R}^{d\times n}\longrightarrow\mathbb{R}^{d\times n}$  consisting of $\frac{1}{\delta}+1$ attention layers, combined with a vector $u\in\mathbb{R}^d$, such that the \textit{sequence ids} $q(Z)\defeq u^{T}g_c(Z)$ satisfies the following properties:
\begin{enumerate}[\hspace{5pt}1.]
    \item For any $\mZ \in {\sZ}_{\delta}$, the entries of $q(\mZ)$ are all distinct.
    \item For any $\mZ, \mZ' \in {\sZ}_{\delta}$ such that $\mZ \neq \mZ'$, all entries of $q(\mZ)$, $q(\mZ')$ are distinct.
\end{enumerate}
\end{lemma}
Such context mappings can map different tokens in the same sequence and same token in different sequence into different ids, which is the key for the transformer to tell different sequences.
To generate such token (column) ids, noting that entries of $\mZ$ are in disjoint intervals separated by $\delta$, we set $\vu=(1,\delta^{-1},\delta^{-2},...,\delta^{-d+1})$. 
Defining $\Delta_{\gamma(i)} = (i-1)\sum_{j=0}^{d-1}\delta^{j}$ for $i>1$ and  $\Delta_{\gamma(1)} = (n-1)\sum_{j=0}^{d-1}\delta^{j}$, then the map $\mZ_{:,i}\to\vu^T\mZ_{:,i}$ from 
$[i-1:\delta:i-\delta]^d$ to $[(i-1)\Delta_{\gamma(j)}:\delta:(i-1)\Delta_{\gamma(j)}+\delta^{-d+1}-\delta]$ is a bijection, and therefore $l_{\gamma(2)}<l_{\gamma(3)}<...<l_{\gamma(n)}<l_{\gamma(1)}$.

\xhdr{Proof of Lemma \ref{aplm:context}}
We first show that the regular sparse self-attention with two heads can implement a \textit{sparse selective shift operation}. For the $i$-th token, the function $\psi(Z;b_Q):\R^{d \times n} \to \R^{1 \times n}$ computes each of its output column as follows:
\begin{equation}
    \psi(\mZ; b_Q)_i
    \defeq \vu^T \mZ_{Ne(i)} \sigma_{H} [(\omega\cdot\vu^T \mZ_{ Ne(i)})^T (\vu^T \mZ_i - b_Q)]=
    \begin{cases}
    \max_{j \in  Ne(i)} \omega\cdot\vu^T \mZ_{j} & \text{ if } \vu^T \mZ_i > b_Q,\\
    \min_{j \in  Ne(i)} \omega\cdot\vu^T \mZ_{j} & \text{ if } \vu^T \mZ_i < b_Q.
    \end{cases}
\end{equation}
By considering two such heads, we have 
\begin{equation}
\label{apeq:headls}
    \Psi(\mZ; c, b_Q, b'_Q) \defeq
    \mZ + 
    \begin{bmatrix}
    c\ve^{(1)} & -c \ve^{(1)}
    \end{bmatrix}
    \begin{bmatrix}
    \psi_l(\mZ; b_Q)\\ \psi_s(\mZ; b'_Q)
    \end{bmatrix},
\end{equation}
where $\ve^{(1)}=[1,0,...,0]$. The $(1,i)$-th entry of $\Psi(\mZ; c, b_Q, b'_Q)$ reads,
\begin{align*}
    \Psi(\mZ; c, b_Q, b'_Q)_{1,i} &=
    \emZ_{1,i} + c(\psi_l(\mZ; b_Q)_i - \psi_s(\mZ; b'_Q)_i)\\
    &=
    \begin{cases}
    \emZ_{1,i} + c(\max_{j \in  Ne(i)} \omega\cdot\vu^T \mZ_{j} - \min_{j \in  Ne(i)} \omega\cdot\vu^T \mZ_{j}) & \text{ if }  b_Q < \vu^T \mZ_i < b'_Q,\\
    \emZ_{1,i} & \text { if } \vu^T \mZ_i \notin [b_Q, b'_Q].
    \end{cases}
\end{align*}
This equation tells us that, if we define the sparse attention layer as $\mZ \to\mZ +  \Psi(\mZ; c, b_Q, b'_Q)$, then any columns $\mZ_{:,i}$ satisfying $\vu^T\mZ_{:,i}\in (b_Q,b'_Q)$ is shifting its first dimension $\mZ_{1,i}$ by the difference between its largest and smallest neighbors, while leaving the remaining column intact.

Next we integrate such sparse attention into the attention diffusion Equation \ref{apeq:attdiff}, and define the token id as $z_i\defeq\vu^T\mZ_{:,i}$.
We consider the attention layer with two head $\psi_l$ and $\psi_s$ as in Equation \ref{apeq:headls}, and focus on the token $\gamma(1)$, whose largest neighbor is itself because of self-connection, and smallest neighbor is $\gamma(2)$. We separate the discussions of two heads $\psi_l$ and $\psi_s$, because $\psi_l$ and $\psi_s$ will make attentions propagate along the larger-token direction (blue arrow in \ref{apfig:gamma}) and smaller-token direction (purple arrows), respectively. 

For the head with $\psi_l$, we define $\omega =\frac{1}{1-\alpha}\delta^{-d}$ and $b_Q<z_{\gamma(1)}$. Then for the first diffusion step, the sparse selective shift will shift the token to its largest neighbor, which is itself: 
\begin{align}
  z_{\gamma(1)}^{(1)} &= (1-\alpha)\omega z_{\gamma(1)}+\alpha z_{\gamma(1)}\nonumber\\
  &=\delta^{-d}z_{\gamma(1)}+\alpha z_{\gamma(1)}\nonumber.
\end{align}

Then for the second diffusion step, we have 
\begin{align}
  z_{\gamma(1)}^{(2)} &= (1-\alpha)\omega z_{\gamma(1)}^{(1)}+\alpha z_{\gamma(1)}\nonumber\\
  &=\delta^{-2d}z_{\gamma(1)}+\alpha\delta^{-d}z_{\gamma(1)}+\alpha z_{\gamma(1)}\nonumber.
\end{align}
By induction, we can see that after the total $K$ diffusion steps, the output of this head will be
\begin{align}
  z_{\gamma(1)}^{(K)} 
  &=\delta^{-Kd}z_{\gamma(1)}+\alpha\sum_{i=1}^{K-1}\delta^{-id}z_{\gamma(1)}+\alpha z_{\gamma(1)}.
\end{align}

Next we consider the head with $\psi_r$, where we define  $\omega =\frac{1}{1-\alpha}\delta^{-d}$ and $b'_Q>z_{\gamma(1)}$, then the diffusion will propagate the attention from token $\gamma(1)$ to $\gamma(2)$, that is,
\begin{align}
  z_{\gamma(1)}^{(1)}
  &=\delta^{-d}z_{\gamma(2)}+\alpha z_{\gamma(1)}\nonumber.
\end{align}
Then for the token $\gamma(2)$, $\gamma(3)$ will be its next smallest neighbors and therefore,
\begin{align}
  z_{\gamma(1)}^{(2)} 
  &=\delta^{-2d}z_{\gamma(3)}+\alpha\delta^{-d}z_{\gamma(2)}+\alpha z_{\gamma(1)}\nonumber.
\end{align}
The diffusion will gradually propagate the attention to the right, and we assume the total diffusion number $K$ is larger than the length of sequence. Therefore, the attention propagation will reach the end token $\gamma(n)$ after $n-1$ steps, and will be stuck at $\gamma(n)$ from then on:
\begin{align}
  z_{\gamma(1)}^{(K)} 
  &=\delta^{-Kd}z_{\gamma(n)}
  + \alpha\sum_{i=n-1}^{K-1}\delta^{-id}z_{\gamma(n)}
  +\alpha\sum_{i=1}^{n-2}\delta^{-id}z_{\gamma(i+1)}
  +\alpha z_{\gamma(1)}.
\end{align}

Next, we combine results from the two heads with $c=1/\alpha$, and calculate the updated token value $\tilde{z}_\gamma(n)$ as

\begin{align}
\tilde{z}_{\gamma(1)} &= z_{\gamma(1)} + \left[\frac{1}{\alpha}\delta^{-Kd}(z_{\gamma(1)}-z_{\gamma(n)}) 
+ \sum_{i=n-1}^{K-1}\delta^{-id}\left(z_{\gamma(1)}-z_{\gamma(n)}\right)
  +\sum_{i=1}^{n-2}\delta^{-id}\left(z_{\gamma(1)}-z_{\gamma(i+1)}\right)
\right].
\end{align}
Then we can see the map from $[z_{\gamma(1)}, z_{\gamma(2)}, ...,z_{\gamma(n)}]$ to $\tilde{z}_{\gamma(n)}$ is one-to-one, because each term $\delta^{-id}$ has different resolution. 
For example, if the $\gamma(1)$-th token in the two sequences has different token ids $z_{\gamma(1)} \neq z'_{\gamma(1)} $, then the first term $ z_{\gamma(1)} - z'_{\gamma(1)}\in[-\delta^{-d+1}+\delta:\delta:\delta^{-d+1}-\delta]$, while the remaining terms have coarser resolution $\delta^{-id+1}$ and cannot cancel the influence of the first term.

Finally, we assume $\gamma(1)$ is chosen as a global token and directly attends to all the remaining tokens and thus we can add an extra single head attention diffusion layer with attention part $\delta^{-(n+1)d-1}\psi(\cdot;0)$, and this operation shifts all tokens by $\delta^{-(n+1)d-1}\tilde{z}_{\gamma(1)}$, which will dominate all remaining token ids.  This ensures that different contexts $\mZ$ are mapped to distinct numbers in $\vu^Tg_c(\mZ)$ thus ensuring the contextual mapping.

\subsubsection{Step 3: Approximating modified Transformer with Diffuser}
Finally, we approximate the modified Diffuser $\overline{\mathcal{D}}$ with the original Diffuser $\mathcal{D}$, by bounding the accumulated errors from $\sigma_H$ and $\phi$ mentioned in Step 2. The detailed process can be found in \citet{yun2019transformers}.

\section{Diffuser as Expander Graphs}
\label{apsec:expander}
\subsection{Equivalent Definitions of Expander Graph}
We define the expansion in the main text from the spectral perspective. Here we show another two commonly-used definitions of expansion.

We first define the the number of edges $E$ between any two subsets $S$ and $T$ of the graph $G=(\mathcal{V},\mathcal{E})$:
\begin{definition}
For $S, T \subset\mathcal{V} $, we denote $E(S,T)=\{(u,v)\in E|u\in S, v\in T\}$.
\end{definition}

Based on this definition, we can then define the edge expansion of $G$ as 
\begin{definition}
The edge expansion ratio of a graph $G$ is 
\begin{equation}
h(G)=min_{S:|S|\leq \frac{|\mathcal{V}|}{2}}\frac{E(S,\bar{S})}{|S|},
\nonumber
\end{equation}
where $\bar{S}$ represents the complement set of $S$ with respect to $\mathcal{V}$.
\end{definition}
Intuitively, this definition tells us how evenly we can divide the graph into two parts, and a graph with small edge expansion has a cut with few crossing edges and divides the vertices approximately in-half.
Similarly, another way to define the expansion is vertex expansion:
\begin{definition}
The vertex expansion ratio of a graph $G$ is 
\begin{equation}
h(G)=min_{S:|S|\leq \frac{|\mathcal{V}|}{2}}\frac{\partial(S)}{|S|},
\nonumber
\end{equation}
where $\partial(S)$ can be either defined as the outer (or inner) boundary of $S$, which is the set of vertices in $\mathcal{V}\setminus S$ (or $S$) with at least one neighbors in $S$ (or $\mathcal{V}\setminus S$).
\end{definition}

Such definitions of expansion can be shown equivalent to the spectral expansion according to Cheeger' Inequality:
\begin{equation}
    \frac{\lambda_2}{2}\leq h(G)\leq \sqrt{2d\lambda_2},
\end{equation}
where $\lambda_2$ is the second smallest eigenvalue of the graph Laplacian.

\subsection {Proof of Theorem \ref{thm:approximate}}
According to Definition \ref{def:approximate}, for two graphs $G$ and $H$,
$(1+\epsilon)H\succeq G \succeq (1-\epsilon)H$ means that 

\begin{equation}
\label{ap_eq:eg1}
\begin{split}
(1+\epsilon)x^T L_H x   \succeq x^T L_G x \succeq (1-\epsilon)x^T L_H x.
\end{split}
\end{equation}
For the complete graph $K_n$, all the eigenvectors $x \perp \vec{1}$ have eigenvalue $n$, i.e,
\begin{equation}
\label{ap_eq:eg2}
\begin{split}
x^T L_{K_n} x   =nx^T  x.
\nonumber
\end{split}
\end{equation}
Let the complete graph we want to estimate to be $H=\frac{d}{n}K_n$, and thus we have
\begin{equation}
\label{ap_eq:eg3}
\begin{split}
x^T L_{H} x   =dx^T  x.
\nonumber
\end{split}
\end{equation}
According to the definition of $(d,\epsilon)$-expander, we have $|d-\lambda_i|\leq\epsilon d$, or $d+\epsilon d \geq \lambda\geq d-\epsilon d$, and therefore
\begin{equation}
\label{ap_eq:eg4}
\begin{split}
(d+\epsilon d)x^T x  \geq \lambda x^T x\geq (d-\epsilon d) x^T x, 
\end{split}
\end{equation}
which is equivalent to \ref{ap_eq:eg1} and proves $G$ is an $\epsilon$-approximation of the complete graph.

\subsection{Proof of Theorem \ref{thm:mixing}}
We assume regular graph $G$ is undirected and unweighted for simplicity. $u$ is the stationary distribution of random walk, which is the uniform distribution in the undirected $d$-regular graph. 
Defining $J=\frac{1}{n}\vec{1}{\vec{1}}^T$, where $\vec{1}$ represents the all-one vector with length $n$, then for any probability vector $p$, we have 
\begin{equation}
\label{ap_eq:jpu}
\begin{split}
\hat{J}p=\frac{1}{n}\vec{1}\vec{1}^T p = \frac{1}{n}\vec{1}\cdot 1 = u.
\end{split}
\end{equation}

Additionally, we observe the eigenvectors of $\hat{A}$ and $A$ are all the same, while the corresponding eigenvalues satisfy $\hat{\mu}_i=\mu_i/d$. Next for the random walk, we have
\begin{equation}
\label{ap_eq:atj}
\begin{split}
    ||\hat{A}^t-J ||_2 &= max_{w:||w ||_2 \leq 1}||(\hat{A}^t-J) ||_2 \\
                       &=\sigma_{max}(\hat{A}^t-J)\\
                       &=\sigma_{max}\left(\sum_{i=1}^{n}\hat{\mu}_i^tv_iv_i^T-\frac{1}{n} \vec{1}\vec{1}^T\right)\\
                       &=\sigma_{max}\left( \sum_{i=2}^{n}\hat{\mu}_i^tv_iv_i^T \right)\\
                       &=max\{|\hat{\mu}^t_2|,|\hat{\mu}^t_n|\}=\beta^t,
\end{split}
\end{equation}
where we use the fact that eigenvector $v_1=\frac{1}{\sqrt{n}}\vec{1}$ and eigenvalue $\hat{\mu}_1=1$. Next,
\begin{equation}
\label{ap_eq:atj1}
\begin{split}
||\hat{A}^tp-u ||_2 &\leq \sqrt{n}||\hat{A}^tp-u ||_2\\
&\leq \sqrt{n}||\hat{A}^tp-Jp ||_2\\
&\leq \sqrt{n}||\hat{A}^tp-J ||_2||p||_2 \leq \sqrt{n}\alpha^t.\\
\end{split}
\end{equation}

The above results can be easily generalized to PageRank where the transition matrix  can be written as $\hat{A} = \alpha\frac{1}{d}A+(1-\alpha)\frac{1}{n}\vec{1}\vec{1}^T$.


\section{Dataset and Experiment Details}
\label{apsec:exp}
For language modeling evaluation, we perform experiments on three tasks including MLM pretraining, long document classification, and QA tasks.  
We use three large language corpora datasets Books \cite{zhu2015book} , CC-News \cite{guu2020retrieval}, and Wikipedia for the model pretraining, with statistics listed in \ref{aptab:preds}. Following RoBERTA, we mask 15\% tokens in the sentence and train the model to predict the mask.We warm start for the first 10,000 steps and then linearly decay the learning rate. The remaining hyperparameters are shown in Table \ref{aptab:prehyper}.

\vspace{-10pt}
\begin{table}[H]
\centering
\begin{tabular}{ccc}
\toprule
Datasets  & \# tokens & Avg. doc. length \\\midrule
Books     & 1.0B      & 37K              \\
CC-News   & 7.4B      & 561              \\
Wikipedia & 3.1B      & 592              \\
\bottomrule
\end{tabular}
\caption{Data statistics of pretraining datasets.}
\label{aptab:preds}
\end{table}
\vspace{-4 pt}
\begin{table}[H]
\centering
\begin{tabular}{ll|ll}
\toprule
Parameters & Values & Parameters & Values \\
\midrule
seq length                & 4096                & local window length             & 64                  \\
\# hidden layers          & 12                  & \# glob tokens                  & 64                  \\
\# heads & 12 & \# rand tokens &64 \\
hidden size               & 768                 & teleport                        & 0.1                 \\
dropout                   & 0.1                 & diffusion steps                 & 5                   \\ \bottomrule
\end{tabular}
\caption{Hyperparameters of pretraining language models. }
\label{aptab:prehyper}
\end{table}
\vspace{-10pt}

For text classification tasks, we use three publicly available datasets, and to evaluate the model's capacity of modeling long sequences, we further introduce two new datasets based on Amazon Product Review. The detailed statistics is listed in Table \ref{aptab:classification_data_full} with model hyperparameter same as in pretraining setting.

\begin{table}[H]
\centering
\begin{tabular}{lrrrrr} \toprule
           & \textbf{HYP} & \textbf{20NG} & \textbf{IMDB}     & \textbf{A-512} & \textbf{A-2048} \\ \midrule
Mean       & 741.44   & 587.56  & 301.14 & 879.62 &  2,915.03\\
Max        & 5,368          & 144,592       & 3,152     & 17,988     &  14,120\\
Min        & 21            & 37           & 10       & 512   &   2,048   \\
Med.        & 547           & 360          & 225      & 725    &  2,505   \\
95pt. & 2,030        & 1,229         & 771      & 1,696  &  5,216    \\
Total      & 645           & 18,846        & 50,000    & 53,471 &  10,000   \\ \midrule
Class      & 2             & 20           & 2        & 5  & 5       \\ \bottomrule
\end{tabular}
\caption{Full dataset statistics on language modeling - text classification evaluation: Hyperpartisan (HYP), 20NewsGroups (20NG), IMDB, Amazon-512 (A-512) and Amazon-2048 (A-2048). \texttt{Med.} is the median value. \texttt{95pt.} indicates 95th percentile. \texttt{Class} indicates the number of classes.}
\label{aptab:classification_data_full}
\end{table}

For QA tasks, we use WikiHop and TriviaQA datasets with detailed statistics listed in \ref{aptab:qa_data_full}. We adopt RoBERTa tokenizer to tokenize the question, answer candidates and supporting contexts, then concatenate them with special tokens. For prediction, we attach a linear layer to attach a linear layer to the special tokens before each candidate answer. We follow the same training setting as in \citet{beltagy2020longformer}, with linear warmup and learning rate 3e-5.

\begin{table}[H] 
\centering
\begin{tabular}{lrr} \toprule
           & \textbf{WikiHop}     & \textbf{TriviaQA}      \\ \midrule
Mean       & 1,564.48 & 1,1641.95 \\
Max        & 21,257       & 172,239      \\
Min        & 73          & 54          \\
Med        & 1,309        & 8,745        \\
25th pctl. & 762         & 3638        \\
50th pctl. & 1309        & 8745        \\
75th pctl. & 2073        & 16286       \\
95th pctl. & 3,672        & 32,158       \\
\midrule
train      & 43,738       & 61,888       \\
val        & 5,129        & 7,993        \\
test       & -           & 7,701   \\ \bottomrule     
\end{tabular}
\caption{Full dataset statistics on WikiHop and TriviaQA for language modeling - question answering evaluation.}
\label{aptab:qa_data_full}
\end{table}

For image modeling experiments, we chose the following baselines to compare: PixcelCNN \cite{pixcelcnn},PixcelCNN++ \cite{salimans2017pixelcnn}, PixelSNAIL \cite{pixelsnail}, Sp. Transformer \cite{child2019generating}, Parallel Multiscale \cite{reed2017parallel} and SPN \cite{menick2019generating}.
For the Long Range Arena (LRA) benchmark, we compare with the following baselines: vanilla Transformer \cite{attention2017vaswani}, Local Attention (a simple local attention baseline), Linear Trans. \cite{kath2020linear}, Reformer \cite{kitaev2020reformer}, Sparse Trans. \cite{child2019generating}, Sinkhorn Trans. \cite{tay2020sparse}, Linformer \cite{wang2020linformer}, Performer \cite{marcin2021performer}, Synthesizer \cite{tay2021synthesizer}, Combiner \cite{ren2021combiner}, Longformer \cite{beltagy2020longformer} and BigBird \cite{zaheer2020bigbird}.
For CIFAR-10, we use the embedding size of 512. Each layer consists of 8 attention heads with hidden dimension 512, and we stack 8 layers. We use the same setting as in \citet{child2019generating}, and train for 500,000 steps with batch size 16 and learning 1e-3. For ImageNet-64 dataset, we scales up the structure to 20-layer and hidden dimension 768, with cosine scheduling learning rate.

\section{More Ablation Studies}
\label{apsec:ablations}
\begin{figure}[H]
\centering
\includegraphics[width=0.7\linewidth]{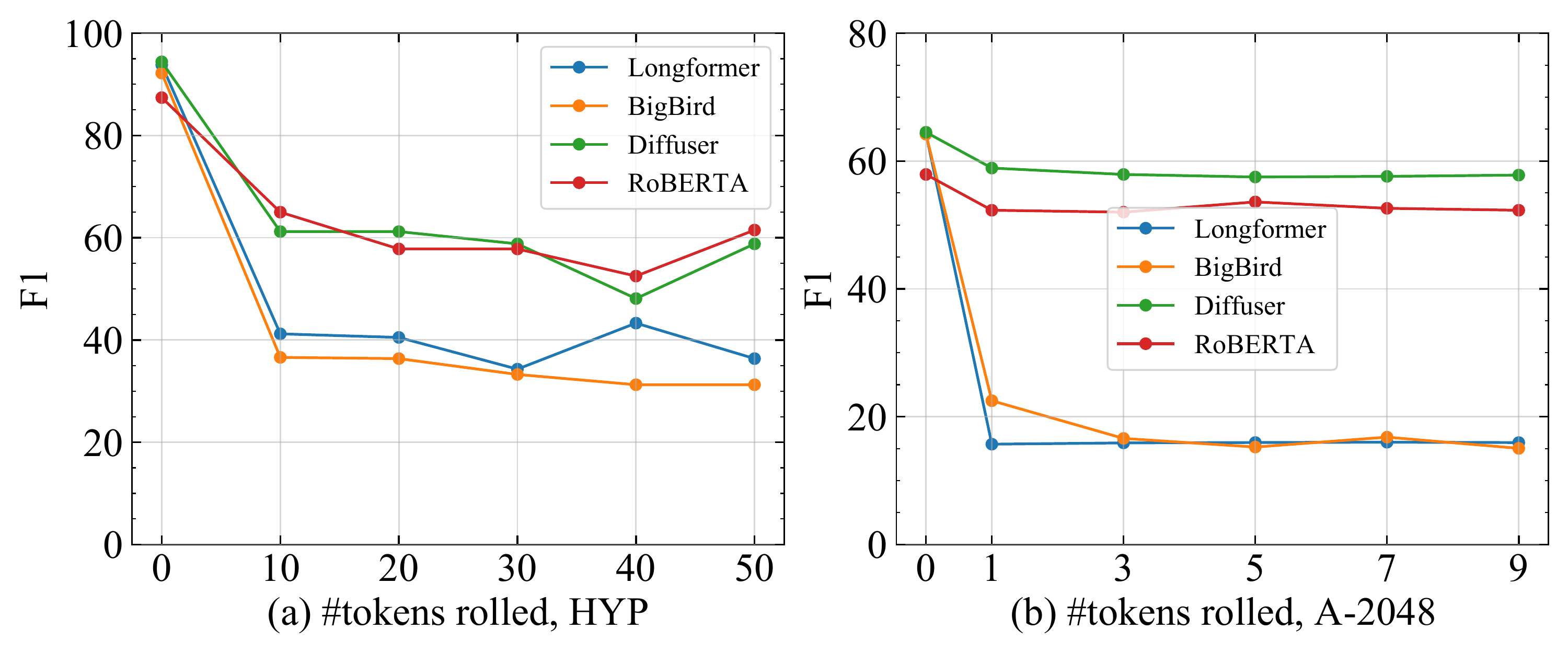}
\caption{Comparison of  the performance drop after rolling the input sequence and mask by few tokens.  }
\label{apfig:roll}
\vspace{-5pt}
\end{figure}

As mentioned in Introduction, one of the challenges for sparse Transformers is the weak robustness to the input sequence changes.  We investigate the influence of perturbing input tokens on the model performance, using HYP (length 4,096) and A-2048 dataset (length 4,096). The input sequence and the corresponding mask are rolled by several tokens and we measure the performance drop. As shown in Figure \ref{apfig:roll}, all models suffer from severe performance degradation even only shifting the input sequences by few tokens (compared to 4096). Such performance drop can be explained by the inconsistency in the attention structure after sequence rolling, and such inconsistency can be further amplified by the changed attention graph for sparse attention. Among different models, Diffuser and RoBERTa are less influenced and therefore have better robustness, which can be explained by the fact that they model the full-attention and the corresponding complete graph will be less affected.

\clearpage

\end{document}